\documentclass[conference]{IEEEtran}
\usepackage{times}
\usepackage{amssymb}
\usepackage{amsmath}
\usepackage{graphicx}
\usepackage[numbers]{natbib}
\usepackage{multicol}
\usepackage[bookmarks=true]{hyperref}
\usepackage{booktabs}
\usepackage{xcolor}
\usepackage{pifont}
\usepackage{wrapfig}
\usepackage{caption}
\usepackage{booktabs} 
\usepackage{multirow}
\usepackage{subfigure}
\usepackage{hyperref}

\begin{document}

\title{CLOT: Closed-Loop Global Motion Tracking for Whole-Body Humanoid Teleoperation}


\author{
    \IEEEauthorblockN{
        Tengjie Zhu\textsuperscript{1,2,*},
        Guanyu Cai\textsuperscript{1,*},
        Zhaohui Yang\textsuperscript{1,*}, 
        Guanzhu Ren\textsuperscript{1},
        Haohui Xie\textsuperscript{1},
        Junsong Wu\textsuperscript{1}, \\
        ZiRui Wang\textsuperscript{2},
        Jingbo Wang\textsuperscript{2},
        Xiaokang Yang\textsuperscript{1},
        Yao Mu\textsuperscript{1,2, $\dagger$},
        Yichao Yan\textsuperscript{1, $\dagger$} %
    }
    \IEEEauthorblockA{\textsuperscript{1}MoE Key Lab of Artificial Intelligence, AI Institute, Shanghai Jiao Tong University}
    \IEEEauthorblockA{\textsuperscript{2}Shanghai AI Laboratory}
    \IEEEauthorblockA{\textsuperscript{*}Authors with equal contribution, \textsuperscript{$\dagger$}Corresponding author}
}



%

\let\oldtwocolumn\twocolumn
\renewcommand\twocolumn[1][]{%
    \oldtwocolumn[{#1}{
    \begin{flushleft}
           \centering
    \vspace{-9pt}
           
    \includegraphics[clip,trim=0cm 0cm 0cm 0cm,width=1.0\textwidth]{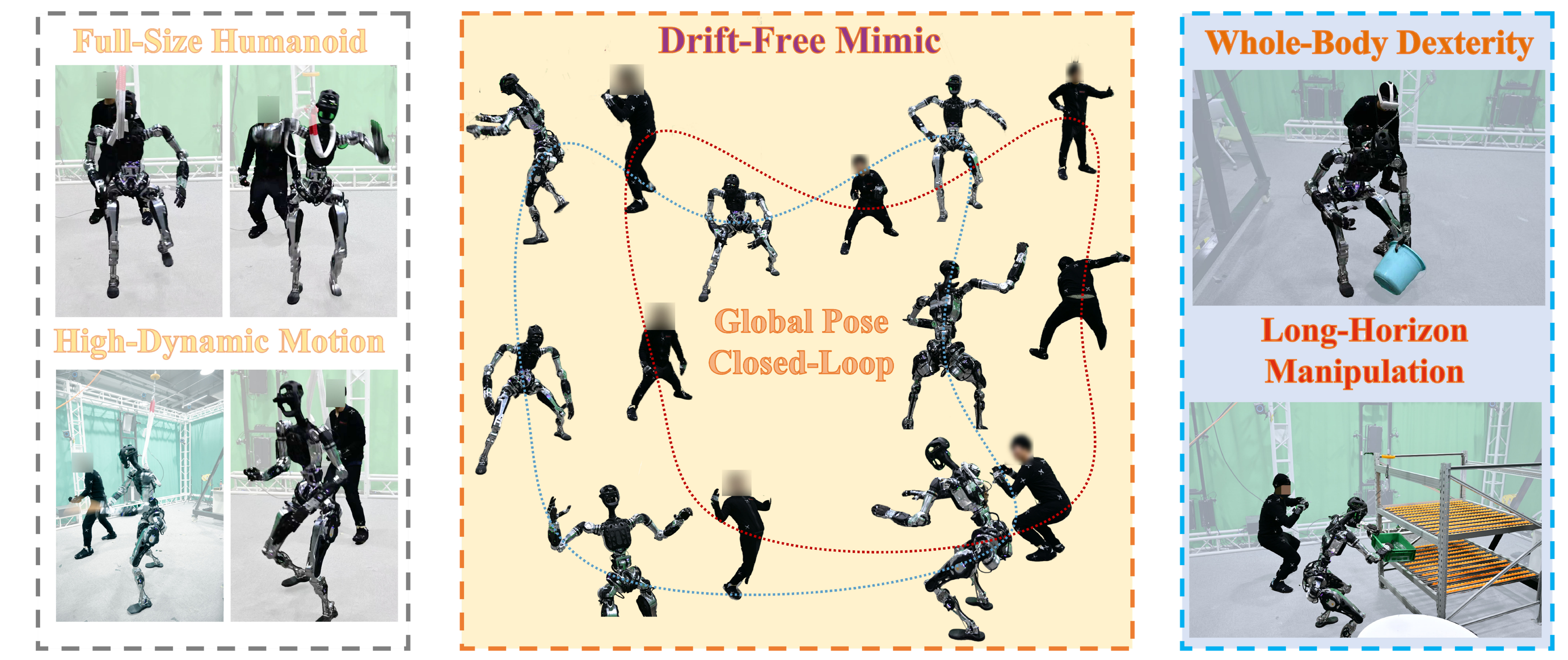}\captionsetup{justification=raggedright,singlelinecheck=false}
    \captionof{figure}{\textbf{Long-horizon whole-body teleoperation with global pose closed-loop feedback.} The proposed framework achieves drift-free human-to-humanoid mimicry on a full-size humanoid, enabling real-world high-dynamic motions and whole-body dexterous manipulation in contact-rich tasks over extended time horizons.}
    \label{fig:teaser}
    \end{flushleft}
    }]
    
}

\maketitle

\begin{abstract}

Long-horizon whole-body humanoid teleoperation remains challenging due to accumulated global pose drift, particularly on full-sized humanoids. Although recent learning-based tracking methods enable agile and coordinated motions, they typically operate in the robot's local frame and neglect global pose feedback, leading to drift and instability during extended execution. In this work, we present CLOT, a real-time whole-body humanoid teleoperation system that achieves closed-loop global motion tracking via high-frequency localization feedback. CLOT synchronizes operator and robot poses in a closed loop, enabling drift-free human-to-humanoid mimicry over long time horizons. However, directly imposing global tracking rewards in reinforcement learning, often results in aggressive and brittle corrections. To address this, we propose a data-driven randomization strategy that decouples observation trajectories from reward evaluation, enabling smooth and stable global corrections. We further regularize the policy with an adversarial motion prior to suppress unnatural behaviors. To support CLOT, we collect 20 hours of carefully curated human motion data for training the humanoid teleoperation policy. We design a transformer-based policy and train it for over 1300 GPU hours. The policy is deployed on a full-sized humanoid with 31 DoF (excluding hands). Both simulation and real-world experiments verify high-dynamic motion, high-precision tracking, and strong robustness in sim-to-real humanoid teleoperation. Motion data, demos and code can be found in our website:
\href{https://zhutengjie.github.io/CLOT.github.io/}{\textcolor{cyan}{CLOT.github.io/}}.

\end{abstract}

\IEEEpeerreviewmaketitle

\section{Introduction}

Collecting high-quality, long-horizon loco-manipulation data is a key prerequisite for advancing general-purpose humanoid robots~\cite{humanoidpolicyhuman}. Robust humanoid teleoperation systems~\cite{teleoperation, humanoid_survey} play a central role in this process by enabling the efficient generation of complex, embodied demonstrations. Ideally, such systems should simultaneously provide (i) whole-body flexibility and coordination required for complex, contact-rich tasks, and (ii) long-term stability to support extended operation without degradation in performance.

Despite substantial progress in whole-body humanoid teleoperation~\cite{mobile, humanplus, homie, amo}, existing systems continue to face a fundamental trade-off between motion agility and long-term stability. Recent real-time whole-body tracking approaches~\cite{twist, twist2} achieve highly flexible joint-level control, enabling expressive and coordinated motions. However, these methods typically operate in the robot’s local frame and neglect global pose feedback. As a result, accumulated global drift becomes unavoidable during extended operation—particularly for full-sized humanoids with higher centers of mass and more challenging control dynamics. Such drift introduces severe safety risks and frequently leads to task failure, as the robot gradually diverges from the operator’s intended global trajectory. While a recent closed-loop teleoperation system~\cite{clone} partially addresses this issue, its control strategy does not support full whole-body tracking, limiting its applicability across diverse teleoperation scenarios.



In this work, we introduce CLOT, a real-time humanoid teleoperation system that enables closed-loop global pose control and global motion tracking via high-frequency localization feedback. We employ an optical motion capture~\cite{optical} system to simultaneously record human motion and the robot’s global pose with high precision. The captured human motions are retargeted online using the Pinocchio~\cite{pinocchio} inverse kinematics (IK) solver to produce robot target trajectories, after which only a general global body tracking controller needs to be trained. However, directly applying existing approaches~\cite{twist, omnih2o} to global pose correction often causes the policy to converge to suboptimal local minima. Specifically, conflicts arise between reward terms for global keypoint tracking and local joint-level objectives, while strongly positive tracking rewards encourage overly rapid convergence, resulting in aggressive and potentially unsafe behaviors.


To address this issue, we propose a new data-driven randomization strategy termed    ``\textbf{Observation Pre-shift}''. In this approach, we periodically set the target pose in the observation to a random future timestamp, while keeping the tracking reward aligned with the current time. This effectively decouples the observation trajectory from the reward trajectory, facilitating the implicit learning of motion interpolation directly from data. 

Learning such implicit interpolation imposes two critical requirements. First, it demands a diverse dataset to ensure generalization. As most open-source datasets~\cite{amass, OMOMO} are primarily designed for animation, a significant portion of these motions exhibit certain discrepancies regarding the kinematic and dynamic properties of physical humanoids.
\begin{wrapfigure}[18]{r}{0.5\columnwidth} 
    \centering
    \vspace{-10pt} 
    \includegraphics[width=\linewidth]{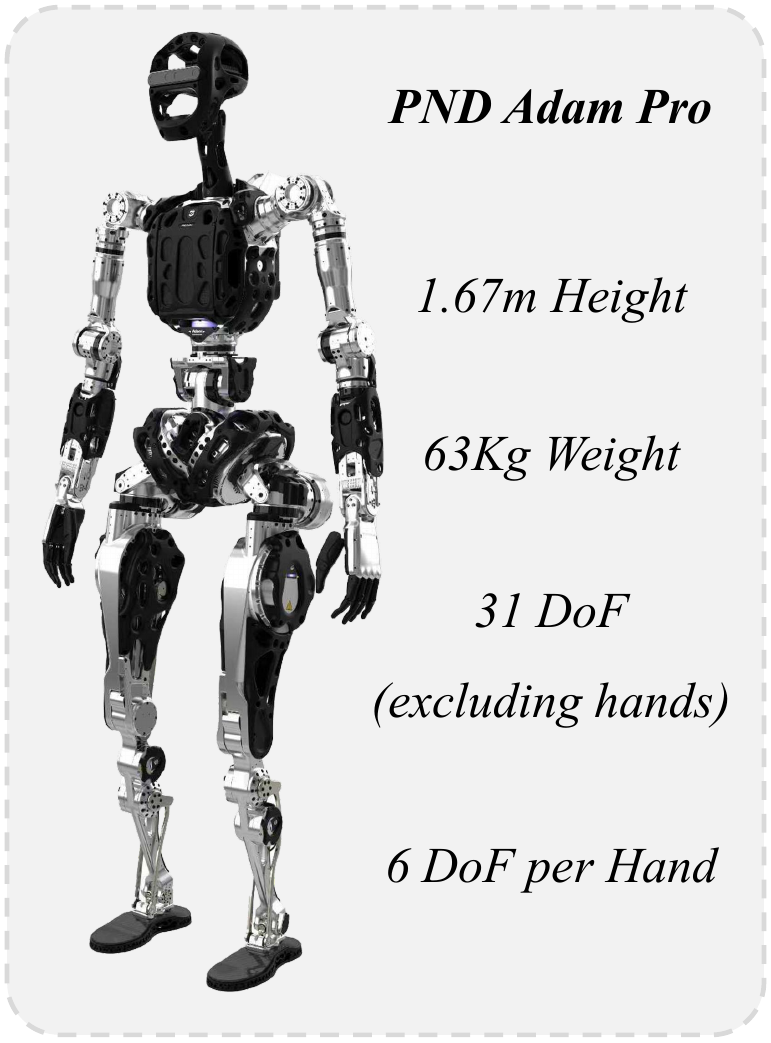}
    \caption{\textbf{Key parameters of PNDbotics Adam Pro.}}
    \label{fig:adam_robot}
    \vspace{-10pt} 
\end{wrapfigure}
To this end, we captured over 20 hours of human data using a strict protocol to preclude unstable behaviors (e.g., tiptoeing, excessive CoM deviation). Furthermore, to enhance the policy network's capability in capturing spatiotemporal information, we designed a Transformer-based~\cite{transformer} architecture. In addition, we incorporate an adversarial motion prior (AMP) reward~\cite{amp} to suppress unnatural motion artifacts.

Finally, we deploy the proposed teleoperation policy on Adam Pro~\ref{fig:adam_robot}, a \textbf{full-sized} humanoid robot with 31 DoF. Experimental results demonstrate that our system maintains long-term stability during extended teleoperation and successfully performs a variety of loco-manipulation tasks. Furthermore, the robot exhibits diverse skills and exceptional disturbance rejection capabilities in real world.


Our main contributions are summarized below:
\begin{itemize}
  \item Closed-Loop Global Control: We introduce CLOT, a real-time whole-body teleoperation system that achieves closed-loop global pose control through localization feedback.
  \item Data-driven randomization strategy: We propose a novel training technique that randomly decouples the policy's observation trajectory from the reward signal, enabling compliant implicit motion interpolation.
  \item Human Motion Dataset: We captured diverse human motions totaling over 20 hours using a strict protocol.
\end{itemize}

\section{Related work}

\subsection{Learning-Based Humanoid Whole-Body Control}

Recent advances in sim-to-real reinforcement learning~\cite{holosoma, schulman2017proximal} (RL) have significantly matured the development of whole-body controllers (WBC) for humanoid robots, enabling a diverse repertoire of agile skills ranging from robust locomotion \cite{loco, HugWBC} to highly dynamic maneuvers such as parkour and dancing \cite{HPL, E2}. By building upon the motion-tracking paradigm \cite{amp, deepmimic}, RL-based frameworks have emerged as a powerful alternative to exhaustive reward engineering. These methods utilize human motion references to synthesize naturalistic and dynamically feasible behaviors, often leveraging high-quality motion retargeting approaches \cite{luo2023perpetual, gmr, omniretarget, pan2025spider, lee2025phuma} to bridge the morphological gap between humans and robots.

Early learning-based trackers focus on robust single-task policy learning using specialized reference motions for gait or athletic behaviors. To improve motion expressiveness and physical feasibility within this paradigm, prior works have incorporated real-to-sim pipelines \cite{he2025asap} and adaptive tracking curricula \cite{PBHC}, enabling the learning of highly dynamic and balance-critical skills \cite{HuB}. Advances in high-performance simulation \cite{mjlab} have further supported efficient and physically faithful sim-to-real transfer for humanoid whole-body tracking. Building on these foundations, recent studies have begun to explore scalable frameworks that learn diverse motions within a unified policy \cite{GMT, UT, any2track, beyondmimic, li2025bfmzero, luo2025sonic}, typically leveraging large-scale public motion datasets \cite{amass, OMOMO, lee2025phuma} to achieve broad motion coverage and tracking robustness. However, existing datasets are not explicitly curated for real-world humanoid teleoperation, whereas our in-house dataset is deliberately optimized for the dynamics of general-purpose humanoids and collected in a task-agnostic manner to better reflect real-world teleoperation conditions.

\subsection{Humanoid Teleoperation Systems}

Humanoid teleoperation is critical for collecting high-quality real-world data necessary to advance embodied intelligence. Prior efforts attempted to address the high-dimensional control challenge by decoupling upper-body manipulation from lower-body locomotion \cite{mobile, deepmimic} via external inputs such as joysticks or exoskeletons \cite{homie, ace}, which ensured stability but inherently restricted the robot's dynamic potential and precluded naturally coordinated behaviors. Recent works have transitioned toward whole-body control strategies, leveraging goal-conditioned tracking optimization \cite{humanplus, omnih2o, amo,  twist, twist2}. These paradigms achieve more fluid and diverse movements, demonstrating an improved capacity to mimic complex human motions on humanoid platforms. However, existing whole-body tracking systems primarily focus on local robot frames, leading to accumulated global pose drift during extended operation. Our work, CLOT, addresses this limitation by incorporating real-time global pose feedback into a whole-body tracking framework to ensure drift-free, long-horizon teleoperation.

\section{Method}
\begin{figure*}[t]
    \centering
    \includegraphics[width=0.98\textwidth]{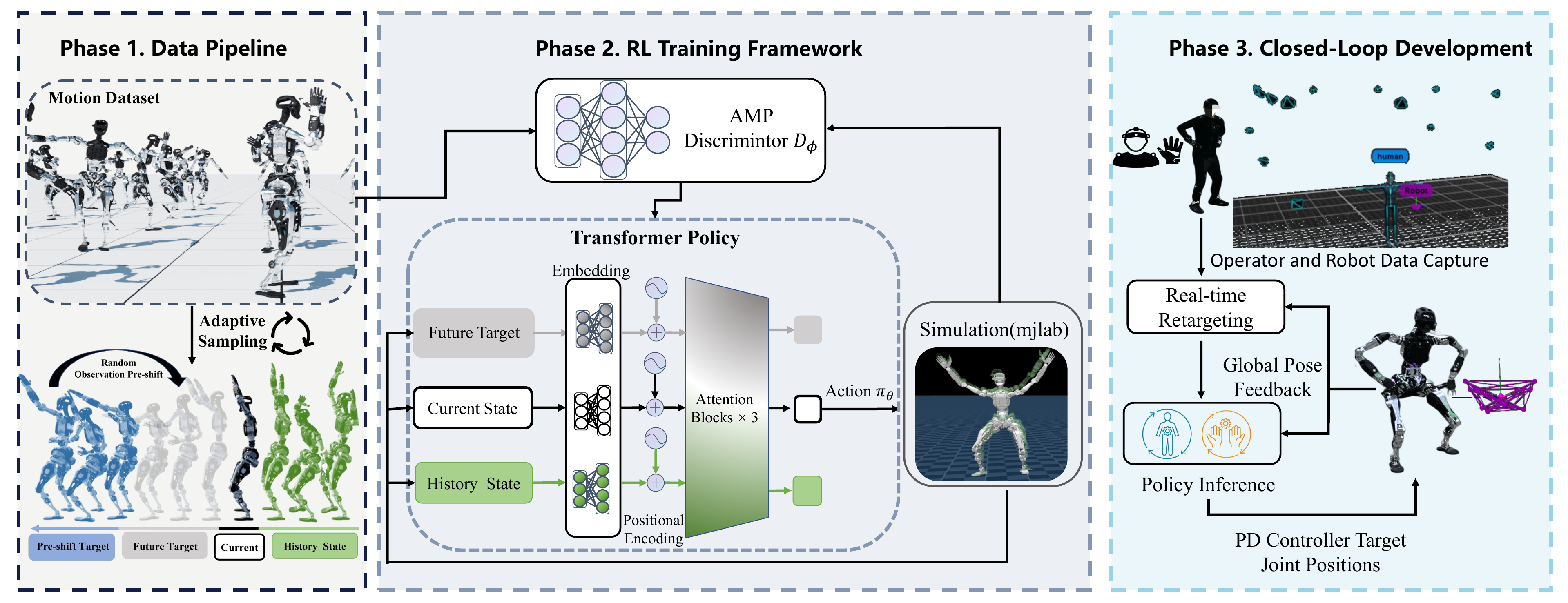}
    \caption{
    \textbf{Overview of the CLOT pipeline.}
    \textbf{Phase 1: Data Pipeline.}
    Human motion is captured using a hybrid optical--inertial motion capture system and retargeted to the humanoid kinematic motion via IK-based retargeting.
    \textbf{Phase 2: RL Training Framework.}
    We propose a random observation pre-shift strategy that decouples policy observations from reward-aligned references, enabling implicit learning of smooth motion interpolation and compliant global corrections. A Transformer policy models long-horizon dependencies, while an adversarial motion prior (AMP) enforces motion realism. Simulation training was performed in mjlab~\cite{mjlab}.
    \textbf{Phase 3: Sim-to-Real Development.}
    Real-time retargeted operator motions provide reference commands, while high-frequency global pose estimates of the robot close the feedback loop, enabling stable and drift-free long-horizon teleoperation.
    }
    \label{fig:pipeline}
\end{figure*}

\subsection{Problem Formulation}

We formulate the humanoid whole-body motion tracking task as a goal-conditioned reinforcement learning (RL) problem. The task is modeled as a finite-horizon Markov Decision Process (MDP), defined by the tuple
$\mathcal{M} = \langle \mathcal{S}, \mathcal{A}, \mathcal{P}, \mathcal{R}, \gamma \rangle$.

At each time step $t$, the agent receives an observation $o_t \in \mathcal{O}$, which consists of the proprioceptive state $s_t \in \mathcal{S}$ and a target motion goal $g_t \in \mathcal{G}$.
The agent outputs an action $a_t \in \mathbb{R}^{23}$, representing proportional-derivative (PD) controller targets for all joints, according to a policy $\pi_\theta(a_t \mid s_t, g_t)$.
The agent receives a scalar reward $r_t = \mathcal{R}(s_t, a_t, g_t)$.
The objective is to learn the optimal policy parameters $\theta$ that maximize the expected discounted return
$\mathbb{E}\!\left[\sum_{t=0}^{T} \gamma^t r_t\right]$.
We train the policy using Proximal Policy Optimization (PPO)~\cite{schulman2017proximal}.

Based on this formulation, our teleoperation system is developed through a pipeline that includes human motion data collection, motion retargeting, RL--based policy training, and the construction of a closed-loop system. For hand and finger motions, we directly apply PD tracking on the retargeted joint references. An overview of the proposed teleoperation pipeline is shown in Fig.~\ref{fig:pipeline}.

\subsection{Human Motion Data Collection}
\begin{figure}[htbp]
    \centering
    \includegraphics[width=0.45\textwidth]{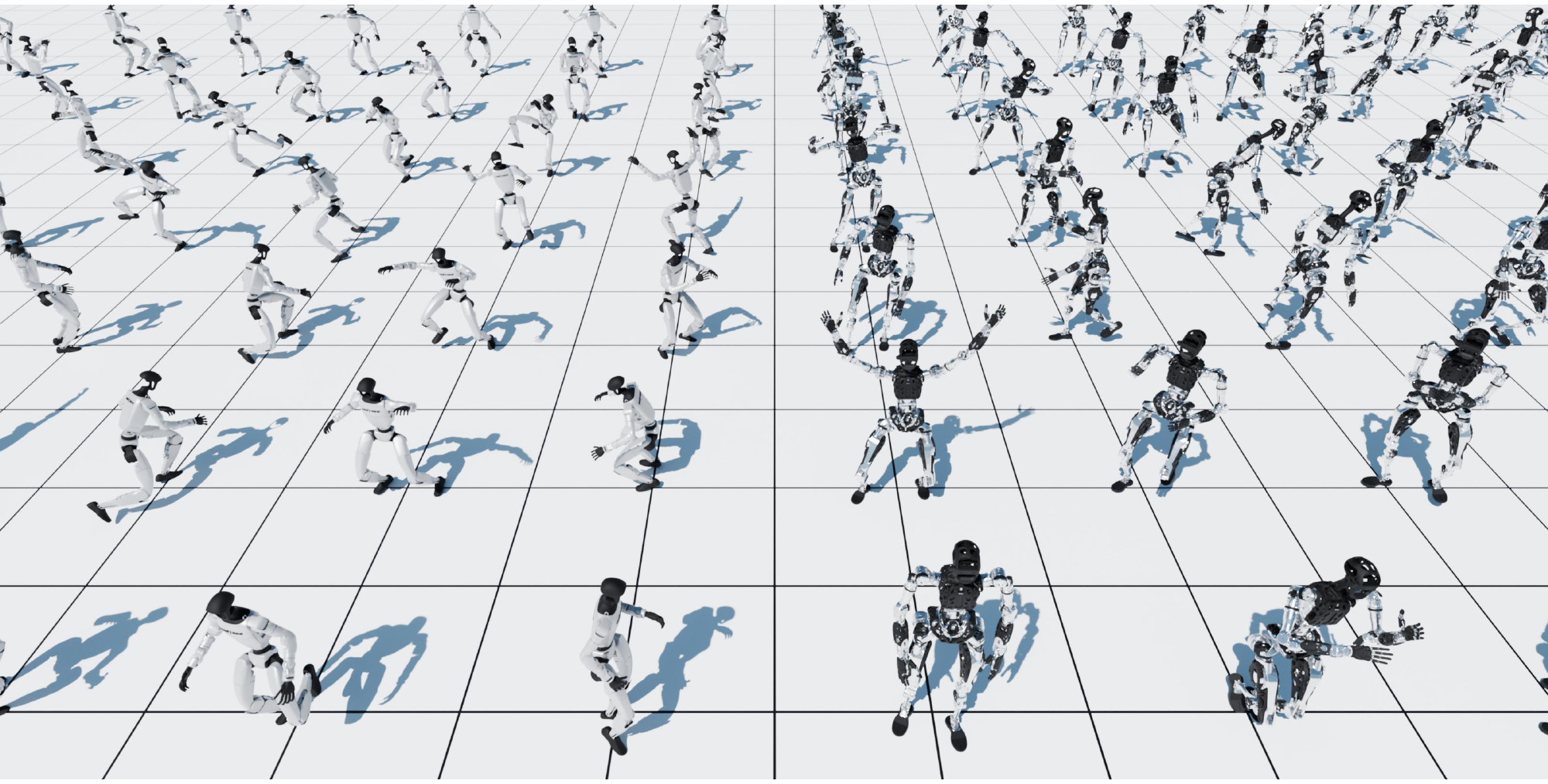}
    \caption{\textbf{Samples from our captured motion dataset.} The dataset includes diverse motions, from locomotion to high-dynamic whole-body behaviors.}
    \label{fig:motion_dataset}
\end{figure}
We collect our own human motion dataset tailored specifically for humanoid robot teleoperation. We employ an OptiTrack optical motion capture system to record approximately 20 hours of high-fidelity full-body human motion data. During data collection, particular emphasis is placed on maintaining physically consistent contacts and motion patterns that are compatible with humanoid kinematic constraints.

Our dataset covers a diverse set of motion categories, with the distribution summarized in Tab.~\ref{tab:motion_distribution}. Representative snapshots are shown in Fig.~\ref{fig:motion_dataset}. The collected motions include both basic locomotion behaviors and high-dynamic whole-body movements, enabling the training of generalizable and robust teleoperation policies. 

\begin{table}[!ht]
\centering
\caption{\centering \textsc{DISTRIBUTION OF COLLECTED HUMAN MOTION DATA ACROSS DIFFERENT MOTION CATEGORIES}}
\label{tab:motion_distribution}
\begin{tabular}{l c l c}
\toprule
\textbf{Motion Category} & \textbf{(\%)} & \textbf{Motion Category} & \textbf{(\%)} \\
\midrule
WALK          & 21.0 & FREE        & 7.0 \\
RUN           & 13.5 & CYCLE       & 5.0 \\
BALANCE       & 10.6 & BOXING      & 4.1 \\
HIGH\_DYNAMIC & 8.9  & CROUCHING   & 3.3 \\
UNLAB         & 9.4  & PICK        & 2.3 \\
SWING         & 9.4  & DANCE       & 2.2 \\
KUNGFU        & 1.7  & LATERAL     & 1.7 \\
\bottomrule
\end{tabular}
\end{table}

\begin{itemize}
    \item \textbf{Basic locomotion patterns:} walking, running, turning, and crouching;
    \item \textbf{High-dynamic whole-body motions:} balancing maneuvers, martial arts techniques, and dance movements.
\end{itemize}

All motions are captured with explicit consideration of foot contact consistency and humanoid actuation limits, ensuring direct usability for whole-body control and reinforcement learning.
\subsection{Real-time Motion Retargeting}
\label{sec:retargeting}
To bridge the gap between human motion and humanoid kinematics, we implement a real-time motion retargeting algorithm.
The primary objective is to minimize tracking errors between robot frames and corresponding human keypoints:
\begin{equation}
    v^* = \arg\min_{v} \sum_{i} w_i \| J_i v - e_i \|^2 + \lambda \| v \|^2,
\end{equation}
where $v$ denotes the joint velocity vector and $e_i$ represents the tracking error for each task. 
We leverage the \texttt{Pinocchio}~\cite{pinocchio} rigid-body dynamics engine to solve high-performance whole-body IK.
To ensure safety, we apply a post-optimization clamping step to the computed velocities and the integrated joint positions.
\subsection{Basic Components for RL Training}

 This section describes the basic components of our RL training framework, including observation design, reward formulation, domain randomization, and curriculum learning.

\paragraph{Asymmetric Actor--Critic Observations}
Following previous works~\cite{he2025asap, PBHC}, we adopt an asymmetric actor--critic observation design. At each timestep $t$, the actor receives $s_t^{\text{actor}} = \big[\, g_{t+1:t+10},\; s_t^{\text{prop}}; s_{t-10:t-1}^{\text{hist}} \,\big]$. The proprioceptive state $s_t^{\text{prop}} =
\big[\,
\boldsymbol{\omega}_t^{\text{root}},\;
\mathbf{g}_t^{\text{proj}},\;
\mathbf{q}_t,\;
\dot{\mathbf{q}}_t,\;
\mathbf{a}_{t-1}
\big]$
includes base angular velocity $\boldsymbol{\omega}_t^{\text{root}}\in\mathbb{R}^3$, projected gravity $\mathbf{g}_t^{\text{proj}}\in\mathbb{R}^3$, joint positions $\mathbf{q}_t\in\mathbb{R}^{23}$, joint velocities $\dot{\mathbf{q}}_t\in\mathbb{R}^{23}$, and the previous action $\mathbf{a}_{t-1}\in\mathbb{R}^{23}$.
A history buffer of length $H{=}10$ stacks recent proprioceptive states, body-position features, and actions to mitigate partial observability.




\paragraph{Tracking-Oriented Reward Formulation}
We design a tracking-oriented reward to encourage accurate teleoperation while penalizing unsafe behaviors and constraint violations. Detailed reward definitions and weighting schemes are provided in the supplementary material.

For tracking-related terms, we follow a commonly adopted formulation~\cite{omnih2o,twist}, where the reward decays exponentially with the tracking error $\mathrm{err}$:
\begin{equation}
    r = \exp\!\left(-\frac{\mathrm{err}}{\sigma}\right),
\end{equation}
with $\sigma$ denoting the error tolerance, which is adaptively optimized during training~\cite{PBHC}. This formulation yields smooth gradients and promotes stable optimization throughout learning.

\paragraph{Domain Randomization} To reduce sim-real gap, we randomize physical parameters and observation conditions, including body mass and inertia, joint damping and friction, actuator strength, contact properties, and observation noise.
The randomized parameters and their ranges are summarized in the supplementary material.

\paragraph{Curriculum Learning}
The high-dimensional action space and complex physical constraints of humanoid robots make early-stage policy learning difficult, especially for full-scale humanoids. We therefore adopt a curriculum learning~\cite{bengio2009curriculum} strategy. Detailed curriculum settings are provided in the supplementary material.




\subsubsection{Adaptive Sampling}
To improve the generalization capability, we adopt a difficulty-aware motion resampling strategy~\cite{beyondmimic} that accounts for varying motion complexities in the dataset.
More challenging motion instances are sampled more frequently, while a clipping mechanism is applied to ensure sufficient coverage of all motions.

\subsection{Key Design Components for RL Policy Training}
To Train control policy that simultaneously achieve generalization capability and high dynamic performance for full-sized humanoid robots, we implemented the following technical approaches:

\subsubsection{Observation Pre-shift}
To achieve robust real-world performance, the policy must retain sufficient exploration diversity during simulation training. Under global motion tracking objectives, prior work~\cite{omnih2o} typically encourages exploration by introducing random perturbations to the robot’s initial states, which often leads to overly aggressive recovery behaviors driven by rapid error minimization.

We propose a \textbf{Randomized Observation Pre-shift} strategy to promote implicit smooth interpolation across reference trajectories. During training, with a randomly sampled probability within a predefined range, the reference observation window is randomly shifted forward by a temporal offset $\delta \in [0, T_{\text{max\_pre}}]$. The resulting actor observation is defined as:
\begin{equation}
    \tilde{s}_t^{\text{actor}} = \big[\, g_{t+1+\delta : t+10+\delta},\; s_t^{\text{prop}},\; s_{t-10:t-1}^{\text{hist}} \,\big].
\end{equation}
Despite the look-ahead reference in the observation, the tracking reward is consistently evaluated against the original reference target at timestep $t$, ensuring strict trajectory fidelity.

By exposing the policy to temporally mismatched reference targets while enforcing current-step tracking objectives, this mechanism encourages the learning of a smooth transition manifold across time. Consequently, the policy exhibits improved stability and continuity when traversing gaps or discontinuities in reference motion sequences. This strategy is applied exclusively during training and is disabled during policy inference.
\subsubsection{Transformer-Based Actor--Critic Network}



To effectively capture spatiotemporal dependencies in the observations and enable stable implicit interpolation, we parameterize both the actor and critic networks using a Transformer architecture. Each observation component is first linearly embedded into a shared token space, and the resulting token sequence is processed by a Transformer encoder that performs global self-attention across historical states, current proprioceptive feedback, and tracking goals. This actor--critic formulation allows the policy to selectively attend to relevant temporal context and goal information conditioned on the robot’s instantaneous physical state, thereby improving stability and expressiveness in long-horizon control.

\subsubsection{Motion Prior reward}

When the robot deviates from the reference trajectory, handcrafted rewards often induce over-aggressive corrections and excessive joint torques.

To mitigate these issues, we incorporate the AMP framework to regularize policy behaviors toward human-like motion styles via adversarial learning. Specifically, we matches the distribution of generated motions $\mathcal{M}_{\text{amp}} \sim G_\theta(z)$, with $z \sim \mathcal{N}(0, I)$, to the style distribution $\mathcal{P}_{\text{style}}$ of a target motion dataset $\mathcal{D}_{\text{motion}}$. We adopt a composite reward formulation
\begin{align}
    r(s_t) &= \lambda_{\text{style}} r_{\text{amp}} + \lambda_{\text{task}} r_{\text{task}}, \\
    r_{\text{amp}} &= \log D_\phi(s_t),
\end{align}
where $D_\phi$ is an adversarial discriminator trained with
\begin{equation}
\mathcal{L}_D = \mathbb{E}_{\mathcal{D}_{\text{motion}}}[\log D_\phi(s)] + \mathbb{E}_{\pi_\theta}[\log(1 - D_\phi(s))].
\end{equation}

The reward term \( r_{\text{amp}} \) enforces stylistic consistency, while \( r_{\text{task}} \) maintains task-level tracking objectives. The two reward terms are balanced using weighting coefficients \( \lambda_{\text{style}} \) and \( \lambda_{\text{task}} \), enabling stable and realistic whole-body motions.

\subsection{Closed-Loop Real-World Teleoperation System}
\begin{figure}[t]
    \centering
    \includegraphics[width=0.8\linewidth]{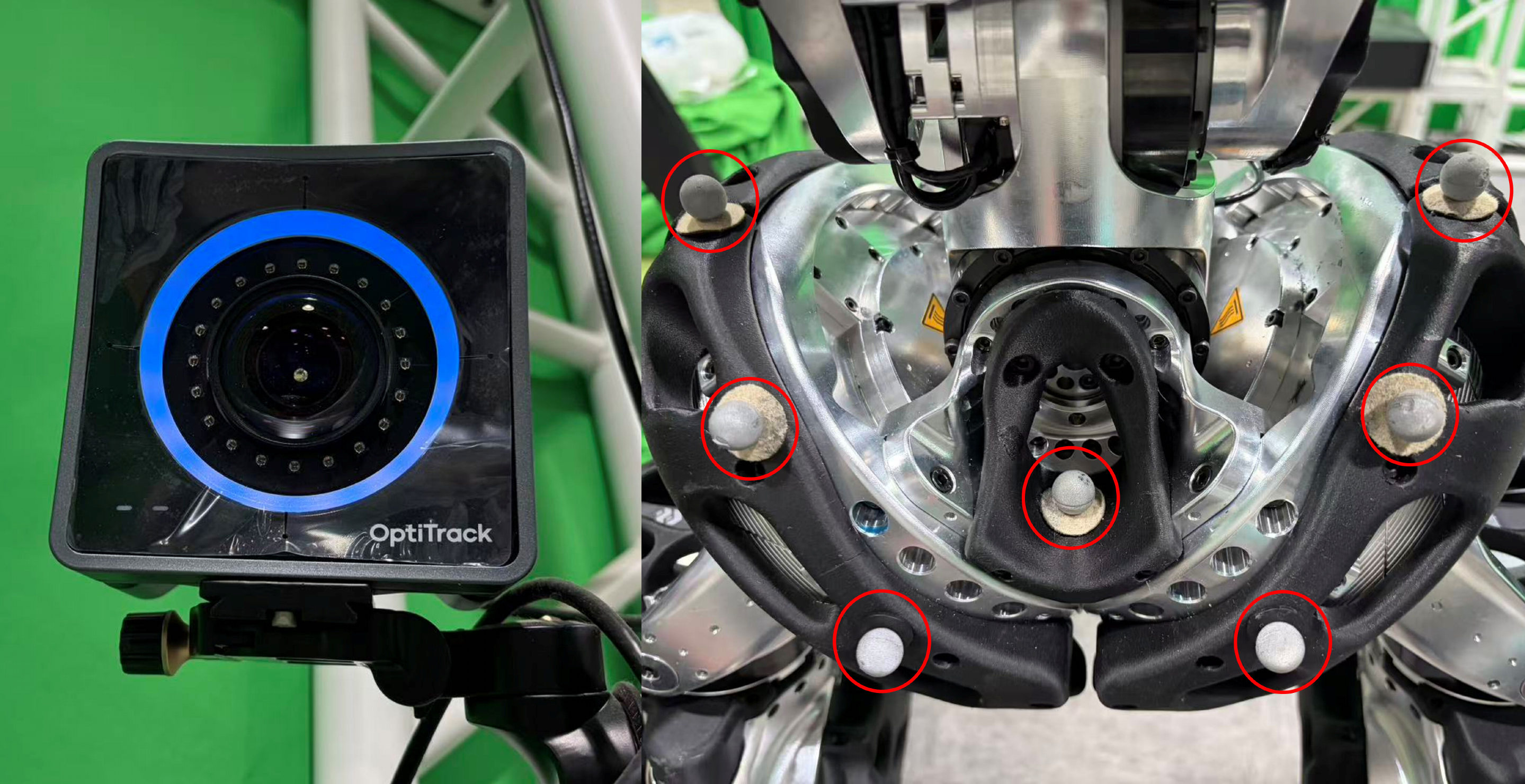}
    \caption{\textbf{Motion capture cameras and marker points.} Our motion capture system obtains the robot’s global pose by tracking the positions of the marker points.}
    \label{fig:OptiTrack}
\end{figure}

Our system establishes closed-loop humanoid teleoperation through real-time synchronization between the human operator and the robot in the global frame. As shown in Fig.~\ref{fig:OptiTrack}, to enable precise and low-latency motion tracking in the real world, we deploy an online streaming pipeline based on a high-fidelity OptiTrack system\cite{optical}, which captures full-body human motion at 120~Hz, along with finger poses recorded at 60~Hz using Hi5 gloves. The captured human motions are streamed online and retargeted to humanoid kinematics using the IK-based retargeting module described in Section~\ref{sec:retargeting}. 

During deployment, the retargeted motion targets are fed our policy, which runs at 50~Hz on an Intel Core i7-14700KF CPU and outputs joint-level position commands. These commands are executed by a low-level PD controller operating at 400~Hz on the humanoid robot. In addition, first-person visual feedback from the robot’s head-mounted camera is streamed to a Meta Quest~3 headset, enabling immersive teleoperation and closed-loop human-in-the-loop control.

\section{Experiments and Results}
\begin{figure*}[htbp]
    \centering
    \includegraphics[width=0.9\linewidth]{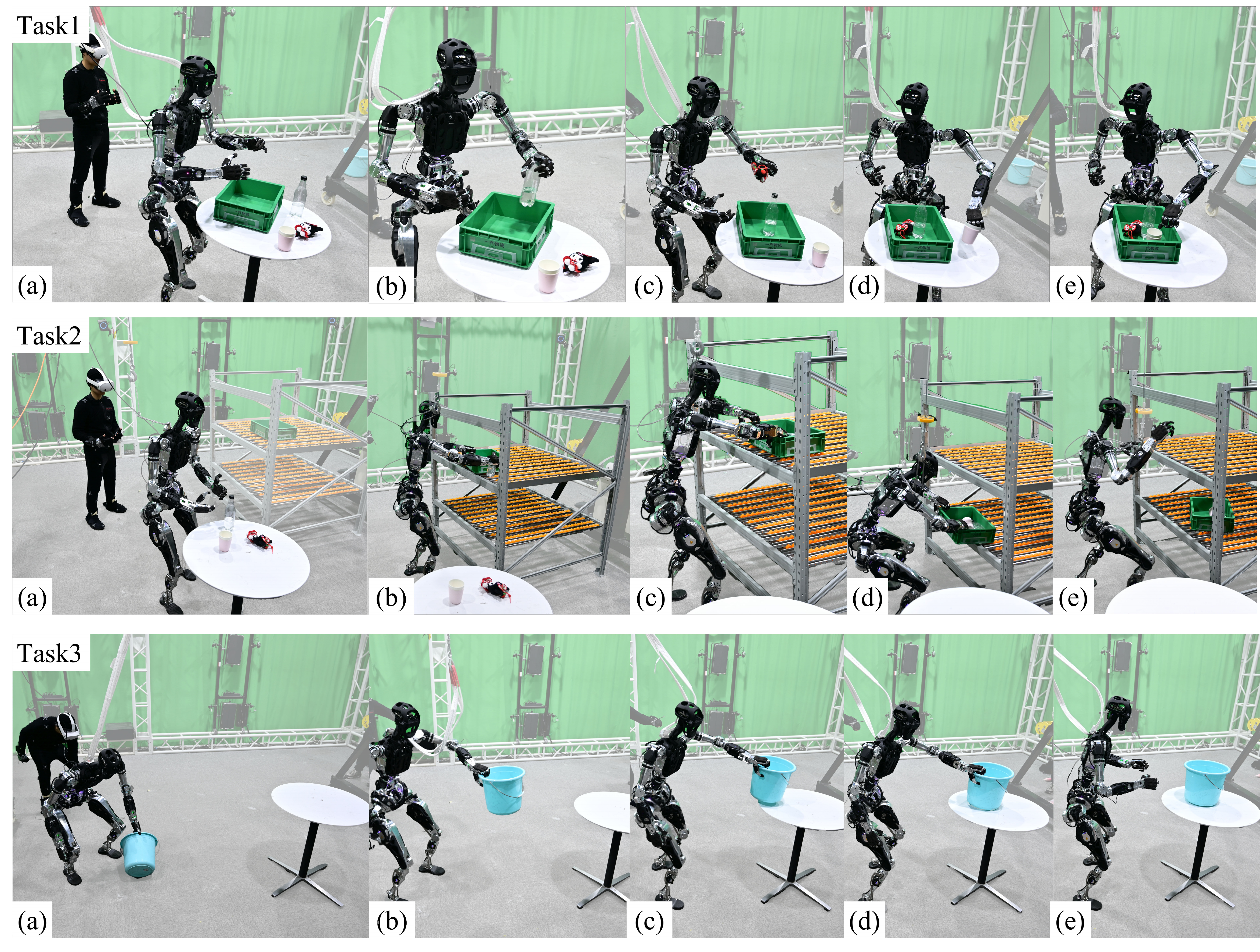}
    \caption{\textbf{Real-world teleoperated loco-manipulation experiments on PNDbotics Adam Pro.} The sequential progression of each task is illustrated from (a) to (e). Throughout the execution, the humanoid robot maintains a constant spatial offset relative to the operator's movements in the world frame. All robot motions across all tasks was governed by a single operator, demonstrating the system's efficiency in whole-body teleoperation.}
    \label{fig:main_expermient}
\end{figure*}
We evaluate our method through extensive simulation and real-world experiments to answer three key questions:
\begin{itemize}
    \item \textbf{Q1:} Does the proposed teleoperation framework generalize across humanoid robots with diverse morphologies and dynamics?

    \item \textbf{Q2:} How do individual components, including observation pre-shift and the Transformer-based policy, contribute to performance and stability?

    \item \textbf{Q3:} Can the full system achieve long-horizon, drift-free, and stable loco-manipulation in real-world tasks?
\end{itemize}

\subsection{Simulation Experiment}
We address \textbf{Q1} and \textbf{Q2} through simulation experiment.

\textbf{Experimental Setting.}
All simulation experiments are conducted in the MuJoCo~\cite{mujoco} environment using two humanoid platforms: Unitree G1 and PNDbotics Adam Pro.
We evaluate our method on a held-out test set consisting of 30 minutes of additional motion capture data.
The test motions cover a diverse set of common human behaviors, including walking, running, dancing, squatting, turning, and jumping.

\textbf{Metrics.} We evaluate tracking accuracy and control smoothness using body-, root-, and joint-level error metrics. Specifically, $E_{mgbp}$ and $E_{mlbp}$ measure global and local body-level tracking errors, $E_{mgrp}$ and $E_{mdp}$ capture root- and joint-level position accuracy, while torque-based metrics reflect control effort and smoothness. $M_{jt}$ and $\sigma_{mjt}$ quantify the magnitude and variability of joint torques, reflecting control effort and smoothness.

\begin{table}[hbp]
\centering
\caption{\textsc{Simulation performance comparison and ablation results across robot platforms.}}
\label{tab:simulation_comparison_all}
\setlength{\tabcolsep}{3pt}
\renewcommand{\arraystretch}{0.95}
\scriptsize
\begin{tabular}{l c c c c c c}
\toprule
\textbf{Method}
& $E_{mgbp}\!\downarrow$ & $E_{mlbp}\!\downarrow$
& $E_{mgrp}\!\downarrow$ & $E_{mdp}\!\downarrow$
& $M_{jt}\!\downarrow$ & $\sigma_{mjt}\!\downarrow$ \\
\midrule

\multicolumn{7}{l}{\textbf{Unitree G1}} \\
\midrule
TWIST2~\cite{twist2}
& 5.094 & 0.340 & 5.164 & 0.364 & \textbf{6.270} & 11.203 \\
\textbf{Ours}
& \textbf{0.056} & \textbf{0.061} & \textbf{0.054} & \textbf{0.158} & 6.604 & \textbf{1.806} \\

\midrule
\multicolumn{7}{l}{\textbf{Adam Pro}} \\
\midrule
Ours (Transf.-only)
& 0.410 & 0.170 & 0.405 & 0.291 & 12.896 & 18.214 \\
\textbf{Ours}
& \textbf{0.116} & \textbf{0.107} & \textbf{0.122} & \textbf{0.136} & \textbf{12.404} & \textbf{8.906} \\

\bottomrule
\end{tabular}
\end{table}

\subsubsection{\textbf{Comparison with Prior Methods}}

To address \textbf{Q1}, we compare our approach against TWIST2~\cite{twist2}, a representative baseline that provides a portable whole-body humanoid teleoperation system for long-horizon operation with full-body pose tracking. TWIST2 is evaluated using its officially released pretrained weights.

Compared with TWIST2, our method reduces all tracking error metrics by more than an order of magnitude on the Unitree G1, while achieving substantially smoother and more stable joint-level control, as summarized in Table~\ref{tab:simulation_comparison_all}. By explicitly incorporating closed-loop global pose feedback, our approach attains particularly strong tracking accuracy.

\subsubsection{\textbf{Ablation Study}}

To address \textbf{Q2}, we conduct ablation studies on the Transformer architecture, the adversarial motion prior reward, and the observation pre-shift mechanism.

Tab.~\ref{tab:simulation_comparison_all} reports the ablation results on PNDbotics Adam Pro.
When only the Transformer network is enabled, the policy exhibits substantially larger tracking errors and higher torque statistics compared to the full system.

These results highlight the importance of incorporating both the motion prior reward and the observation pre-shift mechanism.
The observation pre-shift enables the policy to learn implicit future motion goal differences, which is crucial for avoiding over-aggressive and unstable corrections under large trajectory deviations.

\begin{figure}[htbp]
    \centering
    \begin{minipage}{0.47\linewidth}
        \centering
        \includegraphics[width=\linewidth]{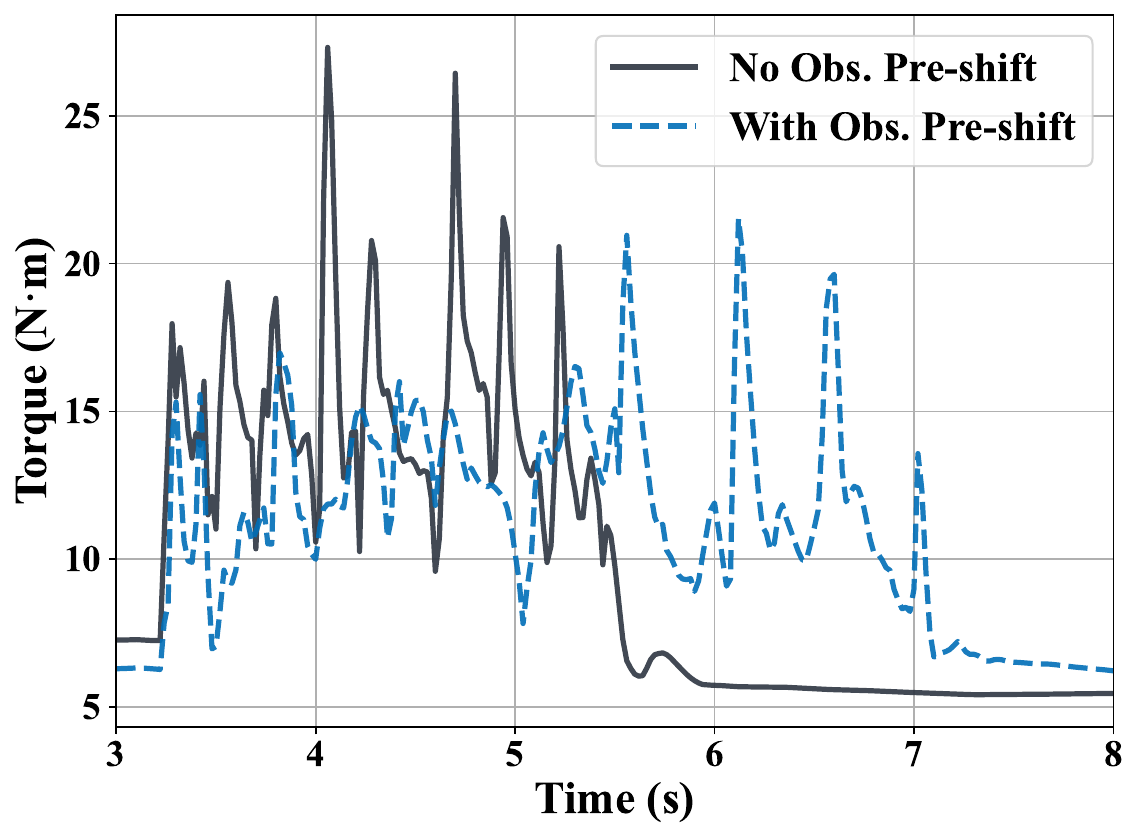}\\
        {\footnotesize (a) AMJT during error recovery.}
    \end{minipage}\hfill
    \begin{minipage}{0.47\linewidth}
        \centering
        \includegraphics[width=\linewidth]{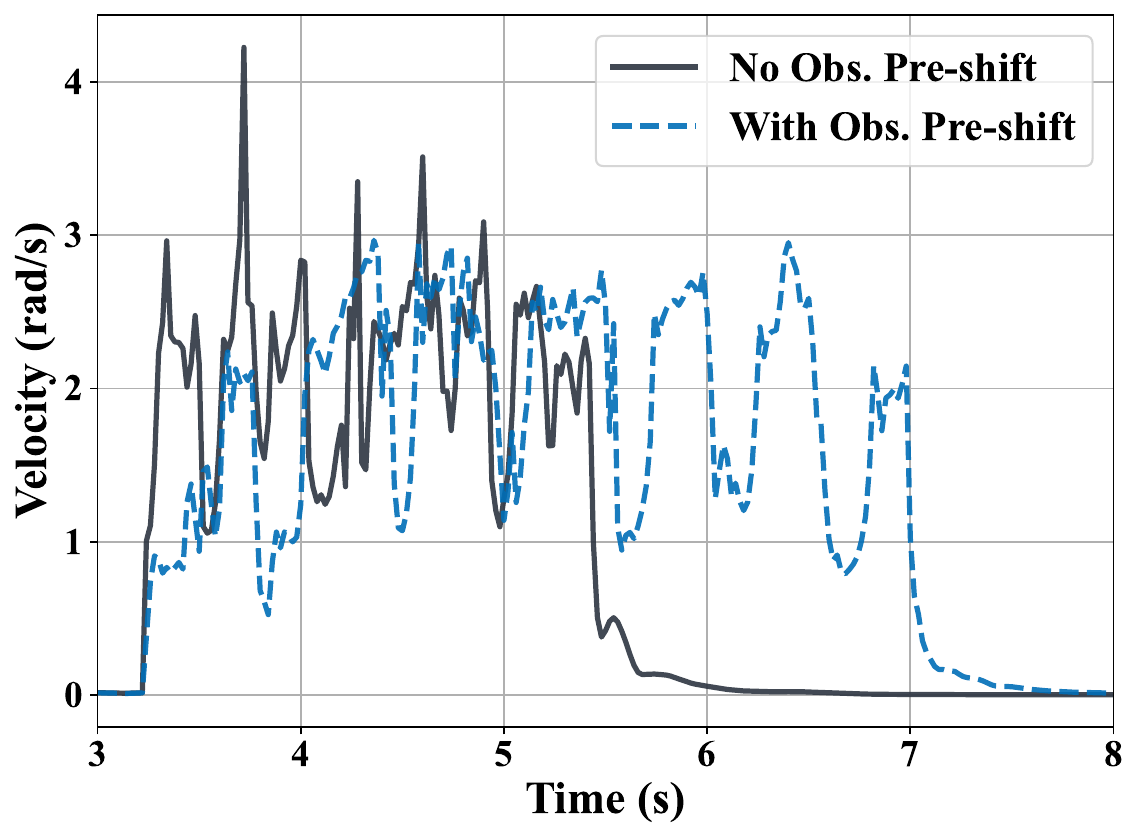}\\
        {\footnotesize (b) AMDV during error recovery.}
    \end{minipage}

    \vspace{0.4em}

    \begin{minipage}{0.47\linewidth}
        \centering
        \includegraphics[width=\linewidth]{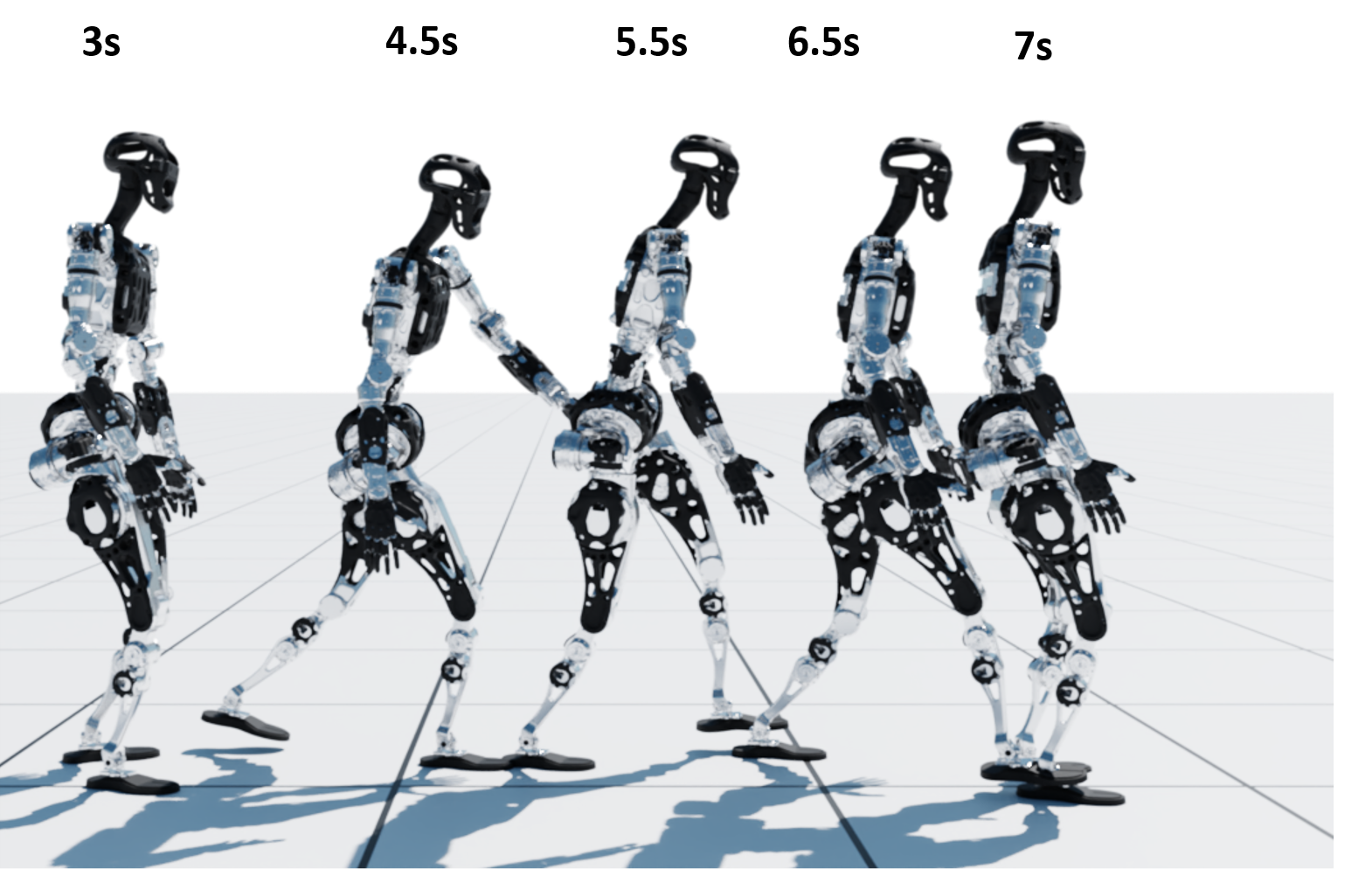}\\
        {\footnotesize (c) Recovery motion sequence with observation pre-shift.}
    \end{minipage}\hfill
    \begin{minipage}{0.47\linewidth}
        \centering
        \includegraphics[width=\linewidth]{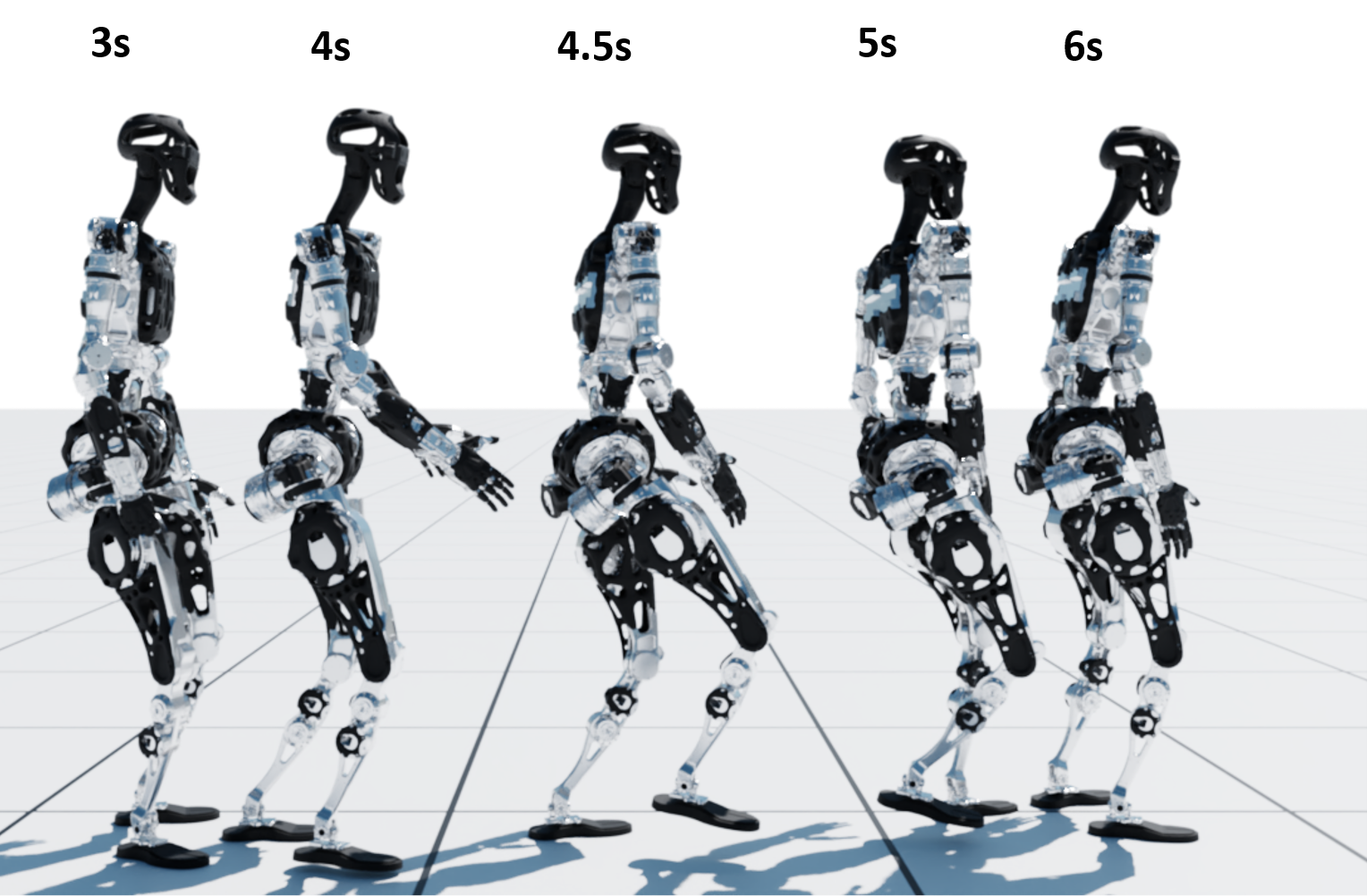}\\
        {\footnotesize (d) Recovery motion sequence without observation pre-shift.}
    \end{minipage}

\caption{\textbf{Ablation study on observation pre-shift mechanism.}
(a,b) Quantitative comparison of AMJT and AMDV during trajectory error recovery.
Policies trained with observation pre-shift exhibit reduced torque and joint velocity peaks.
(c,d) Corresponding recovery motion sequences.
Observation pre-shift enables stable trajectory error correction.}

    \label{fig:adv_compare}
\end{figure}

\begin{table}[h]
\centering
\caption{\centering \textsc{EFFECT OF OBSERVATION PRE-SHIFT ON LARGE-MAGNITUDU TRAJECTORY ERROR CORRECTION}}
\label{tab:advance_comparison}
\begin{tabular}{l l c c}
\toprule
\textbf{Metric} & \textbf{Method} & \textbf{Std$\downarrow$} & \textbf{Smoothness$\downarrow$} \\
\midrule
\multirow{2}{*}{\textbf{Abs. Mean Joint torque}}
 & No Pre-shift   & 1.107 & 7.015 \\
 & With Pre-shift & \textbf{0.990} & \textbf{5.485} \\
\midrule
\multirow{2}{*}{\textbf{Abs. Mean DOF Vel.}}
 & No Pre-shift   & 5.059 & 28.242 \\
 & With Pre-shift & \textbf{3.468} & \textbf{20.442} \\
\bottomrule
\end{tabular}
\end{table}

To further isolate the effect of the observation pre-shift, we simulate abrupt reference motion changes and evaluate the recovery behaviors. As shown in Table~\ref{tab:advance_comparison} and Fig.~\ref{fig:adv_compare}, policies trained with observation pre-shift consistently achieve lower fluctuation and improved smoothness for both absolute mean joint torque (AMJT) and absolute mean DOF velocity (AMDV).

These results indicate that observation pre-shift leads to smoother and more stable trajectory error correction under large reference perturbations.

\begin{figure}[htbp]
    \centering
    \begin{minipage}{0.47\linewidth}
        \centering
        \includegraphics[width=\linewidth]{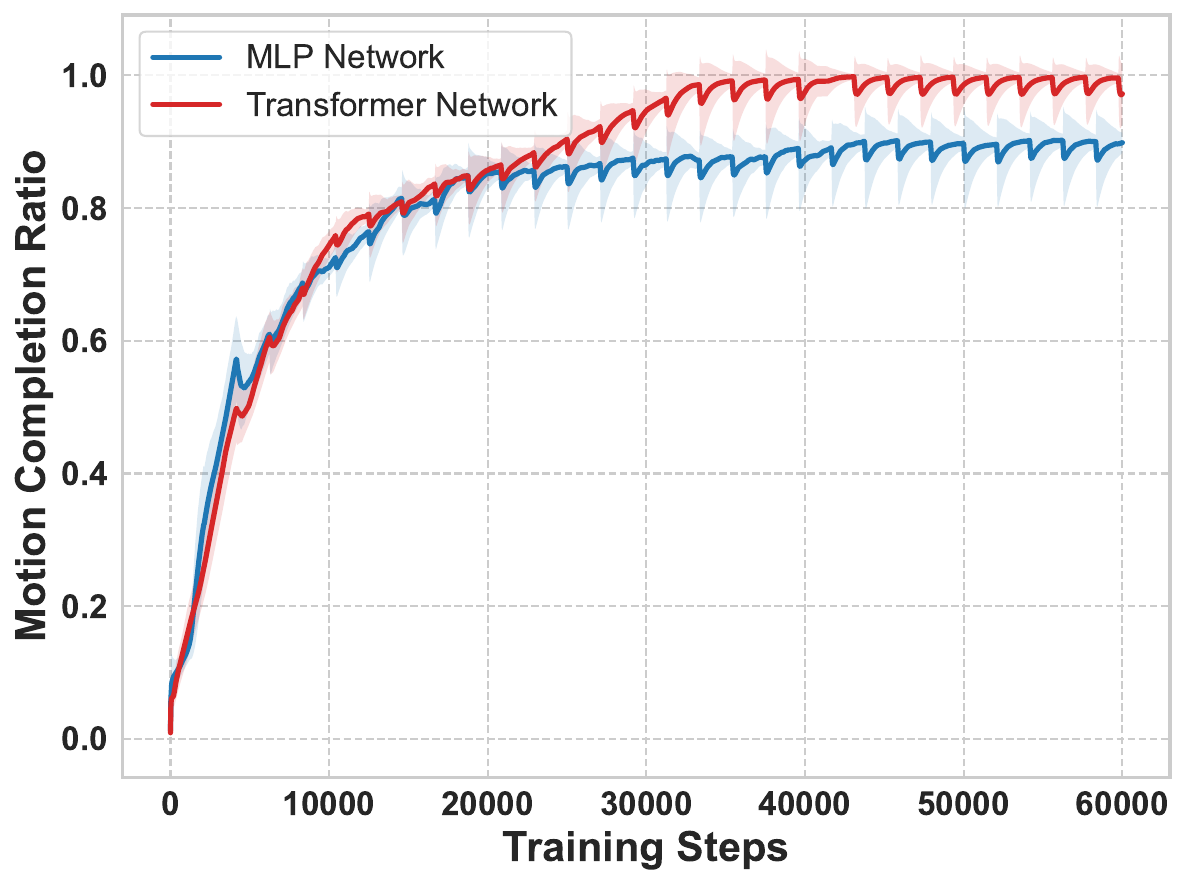}\\
        {\footnotesize (a) Training performance comparison between MLP and Transformer architecture.}
    \end{minipage}\hfill
    \begin{minipage}{0.47\linewidth}
        \centering
        \includegraphics[width=\linewidth]{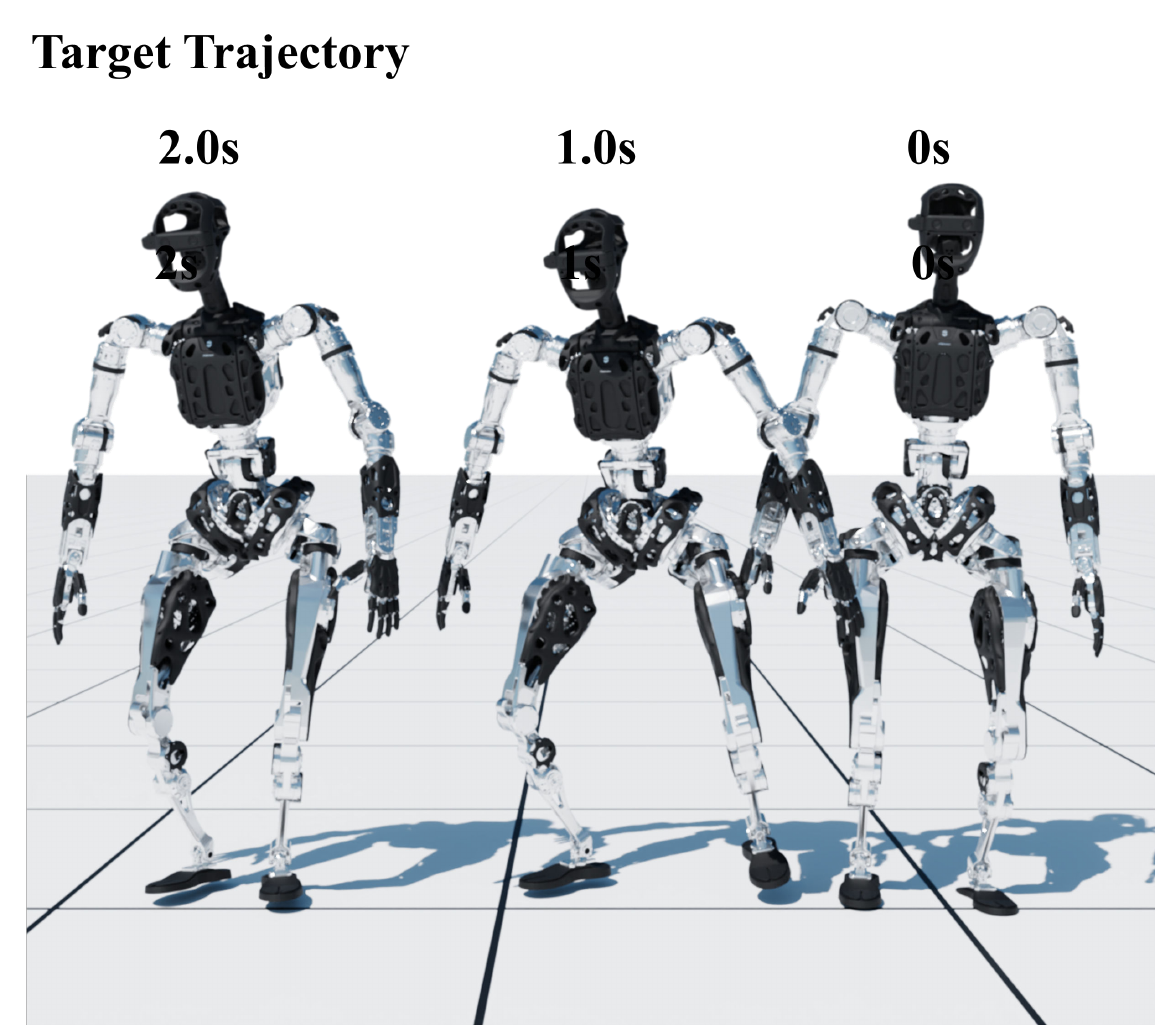}\\
        {\footnotesize (b) Target Trajectory.}
    \end{minipage}
    \vspace{0.4em}

    \begin{minipage}{0.47\linewidth}
        \centering
        \includegraphics[width=\linewidth]{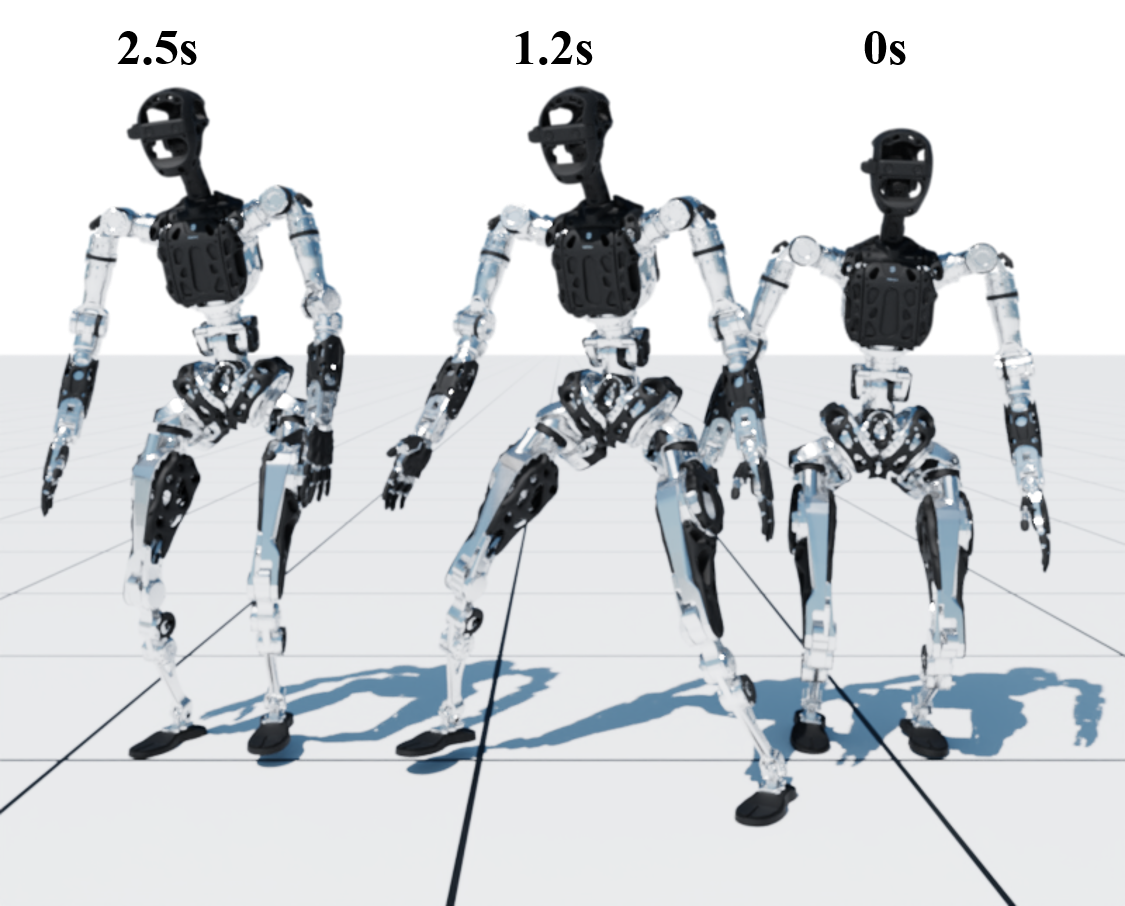}\\
        {\footnotesize (c)Trajectory generated by the Transformer-based policy.}
    \end{minipage}\hfill
    \begin{minipage}{0.47\linewidth}
        \centering
        \includegraphics[width=\linewidth]{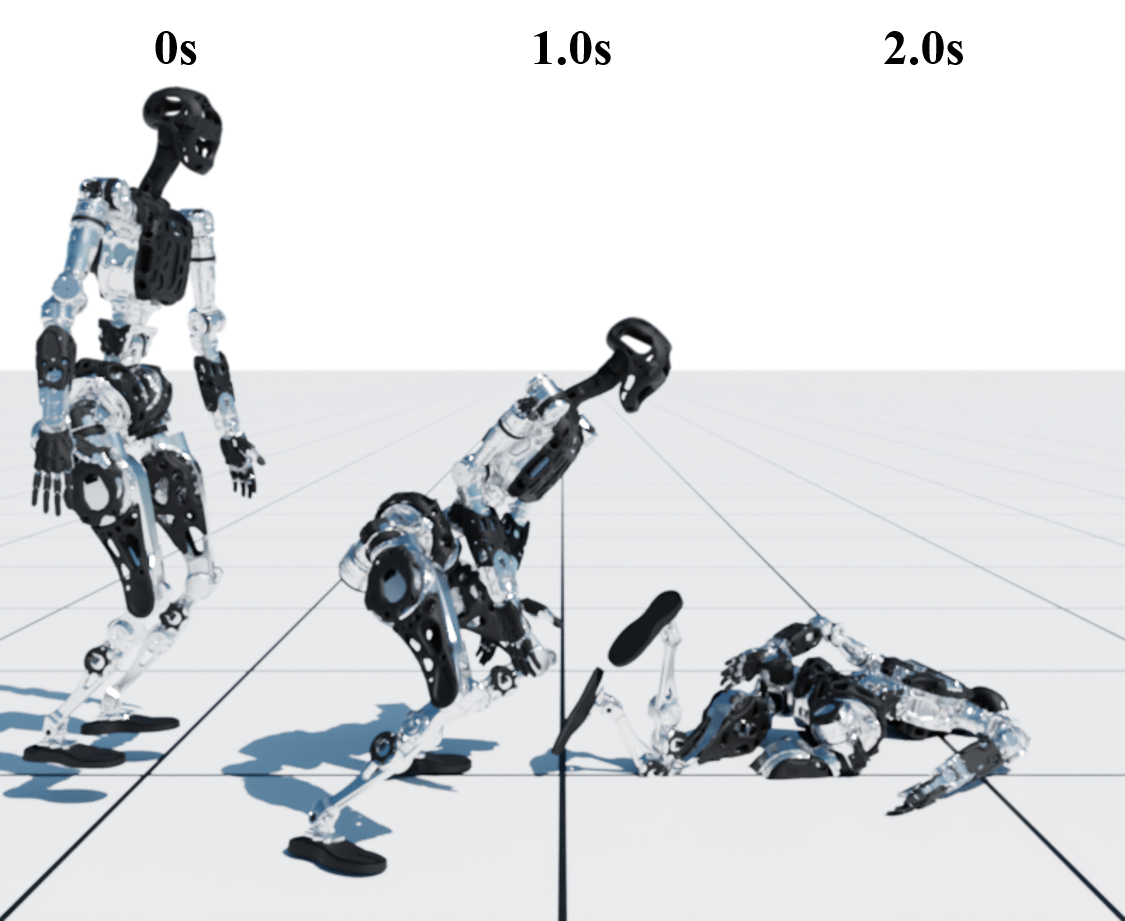}\\
        {\footnotesize (d) Trajectory generated by the MLP-based policy.}
    \end{minipage}

    \caption{\textbf{Network Architecture ablation study.}
The proposed randomization strategy requires capturing long-horizon spatiotemporal dependencies. (a) Training Curve indicate that the MLP fails to converge, whereas the Transformer policy achieves stable learning. (b-d) Trajectory visualization with a pre-shift of 2.0s. The Transformer handles the large offset with stable tracking, whereas the MLP fails and results in a fall.
}
    \label{fig:mlp_transformer_compare}
\end{figure}


Finally, we investigate the interaction between the observation pre-shift mechanism and policy architecture.
Figure~\ref{fig:mlp_transformer_compare} compares the performance of MLP-based and Transformer-based policies under a pre-shift of 2.0\,s.
While the Transformer-based policy successfully captures long-horizon spatiotemporal dependencies and achieves stable tracking, the MLP-based policy fails to converge and results in a fall.
This demonstrates that effectively leveraging observation pre-shift requires sufficient temporal modeling capacity, making the Transformer architecture essential.

\subsection{Real World Teleoperation}
\begin{figure*}[htbp]
    \centering
    \includegraphics[width=0.9\linewidth]{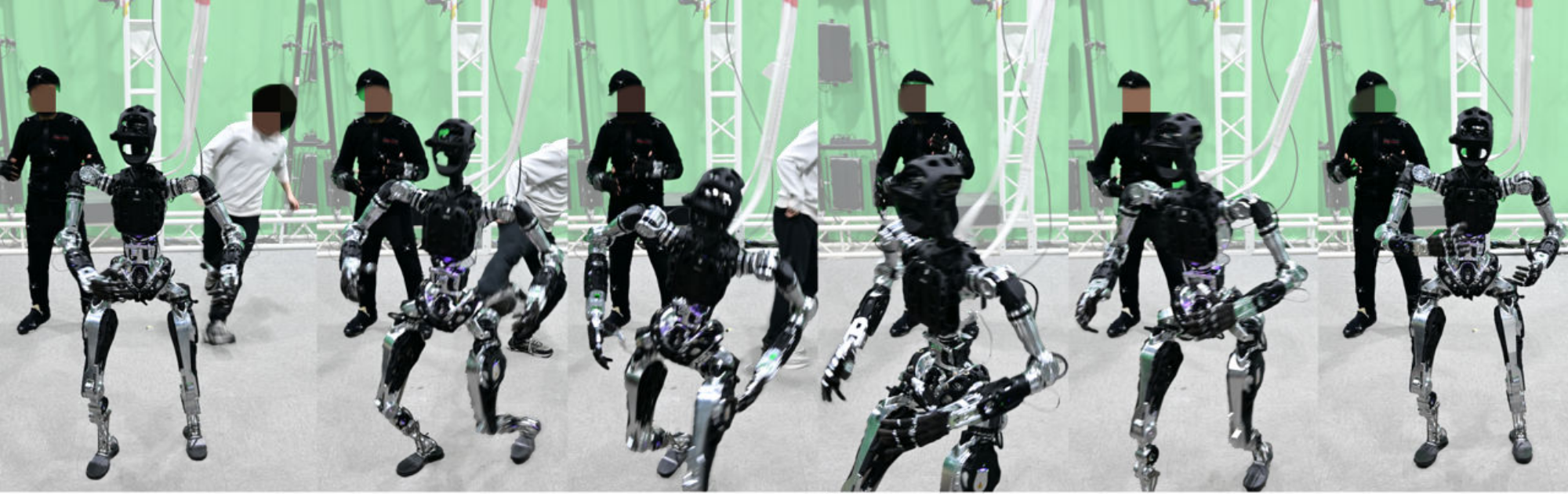}
    \caption{\textbf{Real-world robustness evaluation under strong external disturbances.}
    During teleoperation, the full-sized humanoid robot is subjected to a strong external kick.
    Despite the large impulsive disturbance, the robot successfully recovers its balance and resumes stable motion,
    demonstrating the robustness of our teleoperation framework.}
    \label{fig:real_kick}
\end{figure*}
To answer \textbf{Q3}, we evaluate our system on PNDbotics Adam Pro\cite{PNDbotics}, a full-sized humanoid robot, equipped with dual 6-DoF dexterous hands (see Fig.~\ref{fig:adam_robot}). 
We design three real-world teleoperation experiments: \emph{(i)} long-horizon loco-manipulation teleoperation, \emph{(ii)} extended-duration continuous teleoperation, and \emph{(iii)} robustness evaluation under strong external disturbances.

\subsubsection{\textbf{Long-Horizon Loco-Manipulation Performance}} To evaluate the stability and data collection efficiency of CLOT in long-horizon loco-manipulation tasks, we showcase three representative experiments:
\begin{itemize}
 \item \textbf{Task 1~(T1): Desktop Manipulation.}
The robot performs tabletop pick-and-place with multiple objects, evaluating operational flexibility and control precision against an MPC-based teleoperation baseline.

 \item \textbf{Task 2~(T2): Table-to-Shelf Bin Transfer.} The robot sequentially picks three objects from a table, places them into a top-shelf bin, and then moves the bin to a lower shelf. This long-horizon task challenges whole-body coordination, stable walking, and precise shelf placement.
 \item \textbf{Task 3~(T3): Ground-to-Table Bucket Placement.} The robot retrieves a bucket from the floor and places it upright on a table, requiring low-height grasping, reorientation, and whole-body balance under large CoM shifts.

\end{itemize}

As illustrated in Fig.~\ref{fig:main_expermient}, the robot executes teleoperated loco-manipulation with fluent and coordinated whole-body motion.
We report the success rate (SR) and average task success time (ATST), and compare our method with an MPC-based baseline on the desktop manipulation task in Table~\ref{tab:task_comparison}.

The recorded data highlights our method completes loco-manipulation tasks with high speed and accuracy. Furthermore, it outperforms the MPC approach even in desktop manipulation, thanks to its expanded workspace.

\begin{table}[htbp]
    \centering
    \caption{\centering Performance comparison across real-world teleoperated manipulation tasks.
}
    \label{tab:task_comparison}
    \setlength{\tabcolsep}{4pt} 
    \small
    \begin{tabular}{l l l c c}
        \toprule
        \textbf{Category} & \textbf{Task} & \textbf{Method} & \textbf{SR$\uparrow$} & \textbf{ATST (s)$\downarrow$} \\
        \midrule
        \multirow{2}{*}{Desktop-Manip.} 
            & T1 & MPC-based & 20/30 & \textbf{22} \\
            & T1 & \textbf{Ours} & \textbf{27/30} & 25 \\
        \midrule
        \multirow{2}{*}{Loco-Manip.} 
            & T2 & \textbf{Ours} & \textbf{24/30} & \textbf{82} \\
            & T3 & \textbf{Ours} & \textbf{26/30} & \textbf{42} \\
        \bottomrule
    \end{tabular}
\end{table}

\subsubsection{\textbf{Long-horizon Teleoperation Stability}}
\begin{figure}[h]
    \centering
    \includegraphics[width=0.7\linewidth]{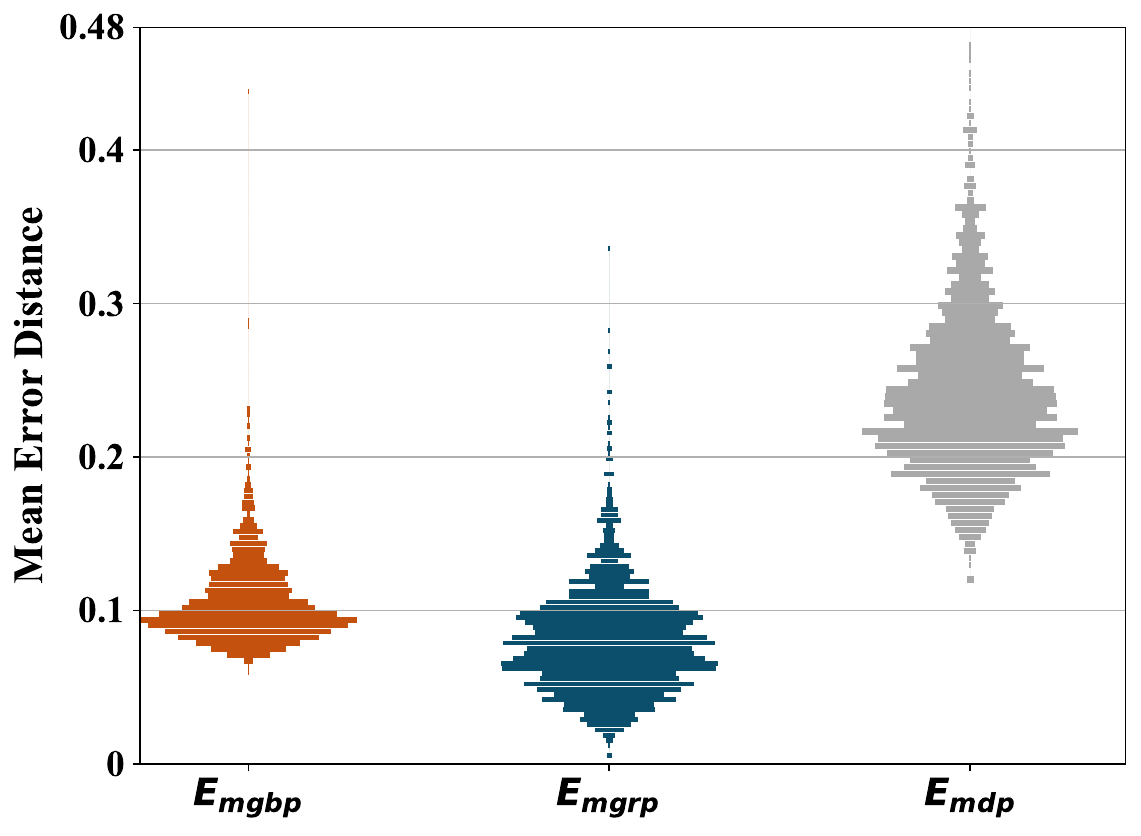}
    \caption{\textbf{Error distributions during long-horizon teleoperated loco-manipulation.}
    Violin plots of the $E_{mgbp}$, $E_{mgrp}$, and $E_{mdp}$ over a 30-minute teleoperation session. The distributions remain stable throughout the long-horizon execution, indicating robust and consistent tracking performance.}
    \label{fig:long_horizon_error}
\end{figure}
To evaluate long-horizon tracking stability, we conduct a 30-minute uninterrupted teleoperated loco-manipulation experiment. Tracking errors are recorded at 1~Hz throughout the session.

All tracking error metrics, including $E_{mgbp}$, $E_{mgrp}$, and $E_{mdp}$, remain consistently low and stable over the entire duration.
The consistently low and bounded error distributions over an extended duration demonstrate the effectiveness of the closed-loop global pose feedback in suppressing drift accumulation, thereby enabling reliable long-horizon teleoperation.



\begin{figure}[h]
    \centering
    \includegraphics[width=1.0\linewidth]{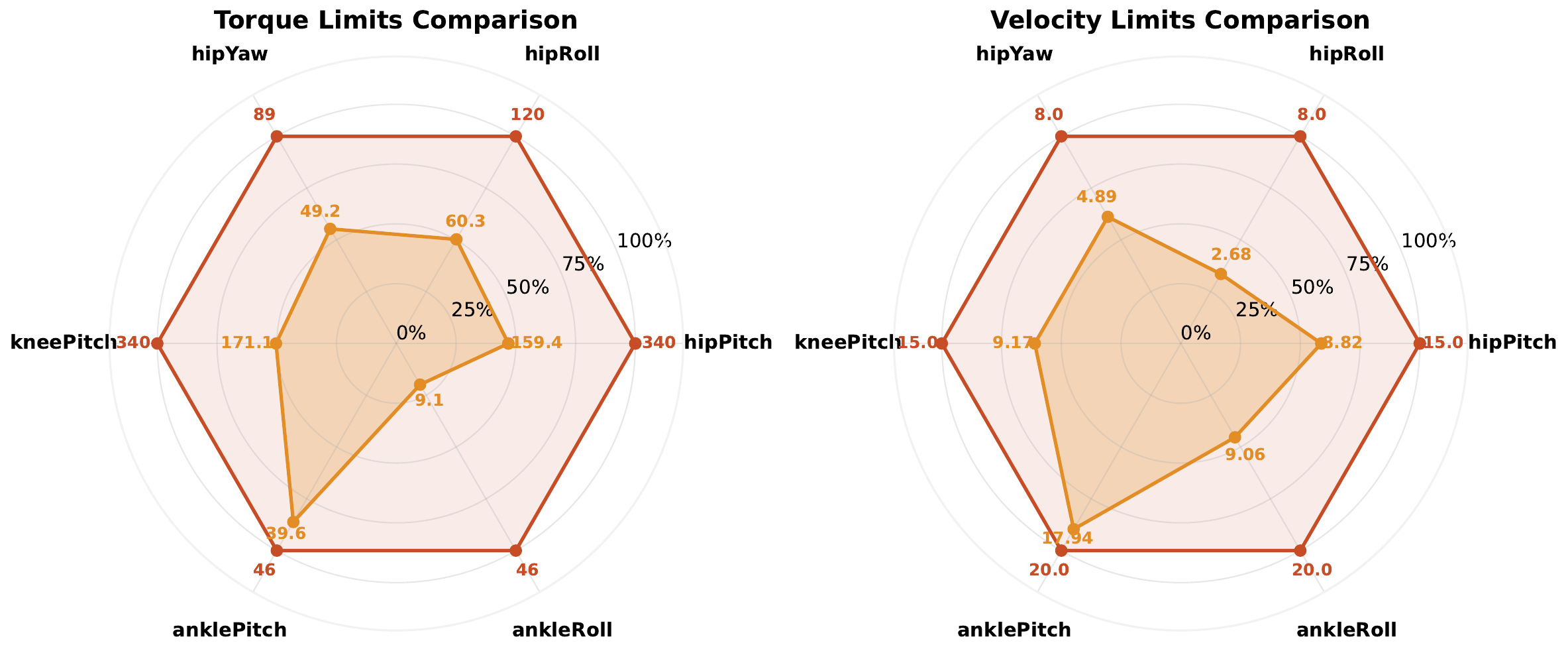}
    \caption{\textbf{Joint-level torque and velocity limits during real-world teleoperation under strong external disturbances.}
Even under a severe external kick, all joints operate within their physical limits.}
    \label{fig:Recovery Torque and Velocity}
\end{figure}

\subsubsection{\textbf{Robustness Performance under Strong External Disturbances}}
To assess robustness under real-world perturbations, we apply strong external kicks to the humanoid robot during teleoperation.
As shown in Fig.~\ref{fig:real_kick}, high-impulse posterior kicks are delivered to the robot. 

Despite sudden perturbations, the robot maintains balance and smoothly resumes its motion trajectory. As shown in Fig.~\ref{fig:Recovery Torque and Velocity}, joint torques and velocities remain within physical limits during impact and recovery, demonstrating robust teleoperation under strong external disturbances in real-world deployment.

\noindent

\section{Conclusion } 
\label{sec:conclusion}



In this work, we presented CLOT, a closed-loop whole-body teleoperation system validated on a full-sized humanoid robot.
We introduce a data-driven randomization strategy to improve stability and smoothness during global error correction, and collect a 20-hour human motion dataset tailored for full-sized humanoid teleoperation. A Transformer-based policy is designed to effectively exploit spatiotemporal information. Together, these components enable robust, high-fidelity, and drift-free teleoperation in real-world deployment. Future work will explore lightweight, cost-effective global localization~\cite{wang,li2024rd} systems and VR-based human motion tracking, to improve system accessibility.

\section{Acknowledgments}

We would like to sincerely thank PNDbotics for providing the robotic platform and comprehensive support related to the robot hardware. We also thank Baidu for providing the GPU resources.



\bibliographystyle{unsrtnat}
\bibliography{references}
\clearpage
\newpage
\begin{appendices}

\section{RL Training Details}
\subsection{Reward Terms and Weights in RL Training}
The reward terms are grouped into positive rewards and penalty terms, as summarized in Table~\ref{tab:reward_weights}. Positive rewards focus on tracking performance at multiple levels, including extended body pose, joint kinematics, keypoint alignment, and foot motion, while also encouraging stable contacts and sustained locomotion through air-time and alive bonuses. Penalty terms are introduced to regularize control behaviors and enforce physical constraints, discouraging excessive torques, abrupt actions, contact instabilities, joint limit violations, and unsafe collisions.

\begin{table}[ht]
\centering
\caption{\centering \textsc{REWARD TERMS AND THEIR WEIGHTS, GROUPED INTO POSITIVE REWARDS ADN PENALTIES}}
\label{tab:reward_weights}
\begin{tabular}{ll}
\toprule
\textbf{Reward term} & \textbf{Weight} \\
\midrule
\multicolumn{2}{c}{\textbf{Positive rewards}} \\
\midrule
Contact Consistency Tracking & 0.5 \\
Joint Limit Satisfaction Reward & 1.0 \\
Extended Body Position Tracking & 1.2 \\
VR Keypoint (3-Point) Tracking & 1.6 \\
Feet Position Tracking & 1.5 \\
Extended Body Orientation Tracking & 1.5 \\
Extended Body Angular Velocity Tracking & 0.6 \\
Extended Body Linear Velocity Tracking & 0.6 \\
Joint Position Tracking & 1.0 \\
Joint Velocity Tracking & 1.0 \\
Feet Air-Time Reward & 160.0 \\
Alive Bonus & 0.2 \\
\midrule
\multicolumn{2}{c}{\textbf{Penalty rewards}} \\
\midrule
Feet Height Deviation Penalty & -20.0 \\
Torque Limit Violation Penalty & -1.0e{-4} \\
Action Rate Penalty & -0.1 \\
Action Smoothness Penalty & -0.2 \\
Feet Contact Force Penalty & -5.0e{-4} \\
Stumbling Penalty & -20.0 \\
Feet Slippage Penalty & -2.0 \\
Joint Position Limit Penalty & -10.0 \\
Joint Velocity Limit Penalty & -10.0 \\
Torque Limit Penalty & -10.0 \\
Collision Penalty & -30.0 \\
Early Termination Penalty & -200.0 \\
\bottomrule
\end{tabular}
\end{table}
\subsection{Observation Setup}
To enhance the performance of our control policy, we provide abundant perceptual information.
The complete specification of the observation components, which is identical for both the Unitree G1 and the PNDbotics Adam Pro, is summarized in Table~\ref{tab:observation}.
\begin{table}[ht]
\centering
\caption{\centering \textsc{OBSERVATION DETAILS}}
\label{tab:observation}
\begin{tabular}{ll}
\toprule
\textbf{Observation term} & \textbf{Dimension} \\
\midrule
last action & 1*23 \\
base ang & 1*3 \\
dof pos & 1*23\\
dof vel & 1*23 \\
future dif pos & 10*29*3 \\
future ref pos & 10*29*3 \\
history base ang & 10*1*3 \\
history projected gravity & 10*1*3 \\
history dof pos & 10*1*23 \\
history dof vel & 10*1*23 \\
history dif pos & 10*29*3 \\
history body pos & 10*29*3 \\
history actions & 10*1*23 \\
\bottomrule
\end{tabular}
\end{table}

\subsection{Domain Randomization Paramters in RL Training}
To improve robustness and reduce the sim-to-real gap, we employ domain randomization during training.
The detailed domain randomization parameters of PNDbotics Adam Pro and their corresponding ranges are summarized in Table~\ref{tab:domain_rand_params}.
All parameters are independently sampled from uniform distributions within the specified ranges. 
\begin{table}[h]
\centering
\caption{\centering \textsc{DOMAIN RANDOMIZATION PARAMETERS}}
\label{tab:domain_rand_params}
\begin{tabular}{l c}
\toprule
\textbf{Domain Rand Params} & \textbf{Range} \\
\midrule
Base COM [m] & $[-0.08, 0.08]$ \\
Link COM [m] & $[-0.02, 0.02]$  \\
Link Mass Scale & $[0.90, 1.10]$ \\
Link Inertia Scale & $[0.85, 1.15]$ \\
PD Gain Scale ($K_p$, $K_d$) & $[0.90, 1.10]$ \\
Friction Scale & $[0.80, 1.50]$ \\
\midrule
External Push Velocity XY [m/s] & $[-0.20, 0.20]$ \\
External Push Interval [s] & $[3.00, 6.00]$ \\
\midrule
RFI Limit Scale & $[0.50, 1.50]$ \\
Control Delay [steps] & $[0.00, 2.00]$ \\
\bottomrule
\end{tabular}
\end{table}

\subsection{Curriculum Learning Paramters in RL Training}
In the early training stage, we reduce penalty, noise and randomization, allowing the policy to focus on learning motion tracking behaviors. As tracking performance improves, these parameters are progressively increased, enabling the policy to adapt to more challenging and realistic conditions. The parameter adaptation rules are defined as follows:

\begin{equation}
    \theta_i^{(t+1)} = 
    \begin{cases}
        \theta_i^{(t)} \cdot \left(1 + \sigma_i\right), & \text{if } \eta^{(t)} > \eta_0 \\
        \theta_i^{(t)}, & \text{otherwise}
    \end{cases}
    \quad \forall\, i \in \{1,2,3,...\},
\end{equation}
where $\theta_i^{(t)}$ denotes the Value of $i$-th parameter at iteration $t$, $\sigma_i > 0$ represents a parameter-specific adaptation rate, $\eta^{(t)}$ denotes the motion completion ratio at iteration $t$, and $\eta_0$ is the performance threshold for triggering updates. Corresponding parameters are summarized in Table~\ref{tab:curriculum_learning_parameters}.

\begin{table}[h]
\centering
\caption{\centering \textsc{CURRICULUM LEARNING PARAMETERS}}
\label{tab:curriculum_learning_parameters}
\begin{tabular}{l c c}
\toprule
\textbf{Curriculum Learning Parameters} & \textbf{Sigma} & \textbf{Range} \\
\midrule
Reward Penalty & 3e{-6} & $[0.05,1.0]$\\
Reward Limits & 2.5e{-7} & $[0.9,0.95]$ \\
Push Robots interval & 1.5e{-5} & $[1,50]$\\
Observation Pre-shift & 5e{-6} & $[1,2]$\\
Termination Distance  & 3e{-6} & $[2.5,3.0]$\\
Termination Scale & 1e{-5} & $[0.8, 4.0]$ \\
Noise & 3e{-6} & $[0.05,1.0]$ \\
\bottomrule
\end{tabular}
\end{table}
\subsection{Detailed Neural Network Architecture}
To enhance the generalization of our control policy, we change our backbone network from MLP to Transformer. Meanwhile, the discriminator network still use MLP. The specific parameters are shown in Table~\ref{tab:nn_architecture}. 

\begin{table}[ht]
\centering
\caption{Neural Network Architecture Parameters}
\label{tab:nn_architecture}
\begin{tabular}{lll}
\toprule
\textbf{Component} & \textbf{Actor \& Critic} & \textbf{Discriminator} \\
\midrule
Architecture & Transformer & MLP \\ 
\midrule
Model dimension & 192 & -- \\
Number of heads & 4 & -- \\
Feedforward dimension & 768 & -- \\
Number of blocks & 3 & -- \\
Number of layers & -- & 4 \\
Hidden units & -- & 1024 \\
Activation & ReLU & ReLU \\
Layer normalization & Yes & Yes \\
\bottomrule
\end{tabular}
\end{table}

\subsection {Proportional–Derivative Controller Paramters}
During training and deployment, the RL policy outputs target joint positions, which are tracked by a joint-space proportional--derivative (PD) controller to generate torque commands:
\begin{equation}
\boldsymbol{\tau} = \mathbf{K}_p (\mathbf{q}^{\star} - \mathbf{q}) + \mathbf{K}_d (\dot{\mathbf{q}}^{\star} - \dot{\mathbf{q}}),
\end{equation}
where $\mathbf{q}^{\star}$ and $\dot{\mathbf{q}}^{\star}$ denote the target joint positions and velocities predicted by the policy.

The PD gains are fixed across all experiments and are selected based on joint functionality.
The complete set of PD controller parameters of PNDbotics Adam Pro is reported in Table~\ref{tab:pd_params}.

\begin{table}[ht]
\centering
\caption{\centering \textsc{PD CONTROLLER PARAMETERS}}
\label{tab:pd_params}
\begin{tabular}{l c c}
\toprule
\textbf{Joint DoF} & \textbf{Stiffness $K_p$} & \textbf{Damping $K_d$} \\
\midrule
Hip Pitch & 305.0 & 5.0 \\
Hip Roll & 255.0 & 3.5 \\
Hip Yaw & 255.0 & 3.5 \\
Knee Pitch & 305.0 & 5.0 \\
\midrule
Ankle Pitch & 50.0 & 0.8 \\
Ankle Roll & 30.0 & 0.35 \\
\midrule
Waist Roll & 255.0 & 3.5 \\
Waist Pitch & 305.0 & 5.0 \\
Waist Yaw & 255.0 & 3.5 \\
\midrule
Shoulder Pitch & 40.0 & 1.0 \\
Shoulder Roll & 40.0 & 1.0 \\
Shoulder Yaw & 40.0 & 1.0 \\
Elbow & 40.0 & 1.0 \\
\bottomrule
\end{tabular}
\end{table}

\section{Experimental Details}
\subsection{Experiment Setup}
\textbf{Compute platform:} The Final policy training was performed in the MJLab simulation using eight NVIDIA RTX 4090 GPUs, with a total training time of approximately one week.
\subsection{Evaluation Metrics}

We evaluate both tracking accuracy and control smoothness using body-, root-, and joint-level metrics.
Let $t=1,\dots,T$ denote timesteps.
We use $\mathbf{p}_{i}(t)\in\mathbb{R}^3$ to denote the Cartesian position of the $i$-th tracked body point from the robot, and $\mathbf{p}^{\star}_{i}(t)$ the corresponding reference target.
Let $\mathbf{r}(t)\in\mathbb{R}^3$ and $\mathbf{r}^{\star}(t)$ be the root (base) positions, $\mathbf{q}(t)\in\mathbb{R}^{n}$ and $\mathbf{q}^{\star}(t)$ be the joint positions, and $\boldsymbol{\tau}(t)\in\mathbb{R}^{n}$ be the joint torques.

\paragraph{Body-level tracking errors}
We report global and local body-position errors.
The global mean body-position error is
\begin{equation}
E_{mgbp}=\frac{1}{T}\sum_{t=1}^{T}\frac{1}{N_b}\sum_{i=1}^{N_b}\left\|\mathbf{p}_{i}(t)-\mathbf{p}^{\star}_{i}(t)\right\|_2 ,
\end{equation}
where $N_b$ is the number of tracked body points.
We also compute the local mean body-position error:
\begin{equation}
E_{mlbp}=\frac{1}{T}\sum_{t=1}^{T}\frac{1}{N_b}\sum_{i=1}^{N_b}\left\|\tilde{\mathbf{p}}_{i}(t)-\tilde{\mathbf{p}}^{\star}_{i}(t)\right\|_2 ,
\end{equation}
where $\tilde{\mathbf{p}}_{i}(t)=\mathbf{R}(t)^{\top}\big(\mathbf{p}_{i}(t)-\mathbf{r}(t)\big)$ and
$\tilde{\mathbf{p}}^{\star}_{i}(t)=\mathbf{R}^{\star}(t)^{\top}\big(\mathbf{p}^{\star}_{i}(t)-\mathbf{r}^{\star}(t)\big)$ are expressed in the root frame, with $\mathbf{R}(t)\in SO(3)$ the root orientation.

\paragraph{Root- and joint-level accuracy}
Root-level position accuracy is measured by
\begin{equation}
E_{mgrp}=\frac{1}{T}\sum_{t=1}^{T}\left\|\mathbf{r}(t)-\mathbf{r}^{\star}(t)\right\|_2 .
\end{equation}
Joint-level (DOF) position accuracy is computed as
\begin{equation}
E_{mdp}=\frac{1}{T}\sum_{t=1}^{T}\frac{1}{n_a}\sum_{j=1}^{n}\left|q_{j}(t)-q^{\star}_{j}(t)\right| ,
\end{equation}
where $n_a$ is the number of actuated joints.

\paragraph{Torque-based control metrics}
To quantify control effort, we compute the mean joint-torque magnitude:
\begin{equation}
M_{jt}=\frac{1}{T}\sum_{t=1}^{T}\frac{1}{n_a}\sum_{j=1}^{n}\left|\tau_{j}(t)\right|.
\end{equation}
To measure smoothness/stability, we report the temporal variability of the per-timestep mean torque magnitude:
\begin{equation}
\sigma_{mjt}=\sqrt{\frac{1}{T}\sum_{t=1}^{T}\left(\bar{\tau}(t)-M_{jt}\right)^2},
\quad
\bar{\tau}(t)=\frac{1}{n_a}\sum_{j=1}^{n_a}\left|\tau_{j}(t)\right|.
\end{equation}
A smaller $M_{jt}$ indicates lower control effort, while a smaller $\sigma_{mjt}$ implies reduced torque fluctuations and smoother control.
\end{appendices}
\end{document}